\RequirePackage[2020-02-02]{latexrelease}
\documentclass[twocolumn, switch]{article} 

\usepackage{preprint}

\usepackage{amsmath, amsthm, amssymb, amsfonts}
\DeclareMathOperator*{\argmax}{argmax} 

\usepackage[numbers,square]{natbib}
\usepackage{courier}
\usepackage[utf8]{inputenc}	
\usepackage[T1]{fontenc}	

\usepackage{booktabs} 		
\usepackage{nicefrac}		
\usepackage{microtype}		
\usepackage{lineno}	    	
\usepackage{float}			
\usepackage{array}
\usepackage{ragged2e}
\usepackage{multirow}
\newcolumntype{C}[1]{>{\Centering}p{#1}}

\usepackage{mathtools}
\DeclarePairedDelimiter\ceil{\lceil}{\rceil}

\usepackage{algorithm2e}
\RestyleAlgo{ruled} 
\SetKwComment{Comment}{/* }{ */}

\usepackage{lipsum}		

\usepackage{graphicx}

\usepackage{titlesec}
\titlespacing\section{0pt}{12pt plus 3pt minus 3pt}{1pt plus 1pt minus 1pt}
\titlespacing\subsection{0pt}{10pt plus 3pt minus 3pt}{1pt plus 1pt minus 1pt}
\titlespacing\subsubsection{0pt}{8pt plus 3pt minus 3pt}{1pt plus 1pt minus 1pt}

\usepackage{tikz,xcolor}
\usepackage{xcolor}		
\usepackage[colorlinks = true,
            linkcolor = purple,
            urlcolor  = blue,
            citecolor = cyan,
            anchorcolor = black]{hyperref}	
            
\definecolor{lime}{HTML}{A6CE39}
\DeclareRobustCommand{\orcidicon}{
	\begin{tikzpicture}
	\draw[lime, fill=lime] (0,0) 
	circle [radius=0.16] 
	node[white] {{\fontfamily{qag}\selectfont \tiny ID}};
	\draw[white, fill=white] (-0.0625,0.095) 
	circle [radius=0.007];
	\end{tikzpicture}
	\hspace{-2mm}
}
\foreach \x in {A, ..., Z}{\expandafter\xdef\csname orcid\x\endcsname{\noexpand\href{https://orcid.org/\csname orcidauthor\x\endcsname}
			{\noexpand\orcidicon}}
}

\title{Generating Hierarchical Structures for Improved Time Series Classification Using Stochastic Splitting Functions}

\usepackage{xwatermark}
\newwatermark[firstpage,color=gray!90,angle=0,scale=0.28, xpos=0in,ypos=-5in]{*correspondence: \texttt{celal.alagoz@gmail.com}}

\usepackage{authblk}

\author[1\thanks{\tt{celal.alagoz@gmail.com}}]{Celal Alagoz\orcidA{}}

\affil[1]{Department of Electrical and Electronics Engineering, Kırıkkale University}

\begin{document}

\twocolumn[ 
  \begin{@twocolumnfalse} 

\maketitle
\begin{abstract}
This study introduces a novel hierarchical divisive clustering approach with stochastic splitting functions (SSFs) to enhance classification performance in multi-class datasets through hierarchical classification (HC). The method has the unique capability of generating hierarchy without requiring explicit information, making it suitable for datasets lacking prior knowledge of hierarchy. By systematically dividing classes into two subsets based on their discriminability according to the classifier, the proposed approach constructs a binary tree representation of hierarchical classes. The approach is evaluated on 46 multi-class time series datasets using popular classifiers (\texttt{svm} and \texttt{rocket}) and SSFs (\texttt{potr}, \texttt{srtr}, and \texttt{lsoo}). The results reveal that the approach significantly improves classification performance in approximately half and a third of the datasets when using \texttt{rocket} and \texttt{svm} as the classifier, respectively. The study also explores the relationship between dataset features and HC performance. While the number of classes and flat classification (FC) score show consistent significance, variations are observed with different splitting functions. Overall, the proposed approach presents a promising strategy for enhancing classification by generating hierarchical structure in multi-class time series datasets. Future research directions involve exploring different splitting functions, classifiers, and hierarchy structures, as well as applying the approach to diverse domains beyond time series data. The source code is made openly available to facilitate reproducibility and further exploration of the method.

\end{abstract}
\vspace{0.35cm}
  \end{@twocolumnfalse} 
] 


\section{Introduction}
HC is a method of organizing data or objects into a tree-like structure of nested categories or groups, where each category is a subset of a larger category, forming a hierarchy. HC is typically used in many fields such as text classification \cite{klimt2004enron,punera2005automatically}, image understanding \cite{shen2003multilabel} and annotation \cite{dimitrovski2011hierarchical, dimitrovski2012hierarchical}, and in bioinformatics problems such as protein function prediction \cite{clare2003machine,barutcuoglu2006hierarchical, silla2009global, cesa2009hierarchical} where usually large set of labels are designated with a hierarchical structure in advance. In such cases, algorithms have been developed to take advantage of the structure to improve classification accuracy. However, most multi-class classification problems do not have structured labels, and the potentiality of performance improvements which was seen in hierarchical problems are under explored in those domain of applications. One such field is time series classification where multi-class labels assume traditional flat labels.

Only a limited number of studies have reported the benefits of inducing a hierarchy from datasets that are typically utilized with flat labels. In a web content analysis study, an automatic taxonomy of documents when they are not pre-defined was retrieved via construction of a binary tree  \cite{punera2005automatically}. Tree construction was performed with hierarchical clustering of classes in a top-down approach where cluster splitting was done Spherical K-means. Features to distinguish classes were selected using Fisher index criteria. In a later study \cite{punera2006automatic} they reported superiority of nary tree over binary tree in classification performance. Assuming presence of a latent hierarchy on synthetic and various image data has shown to result in notable improvement in down-stream classification task \cite{helm2021inducing}. In their investigation, two distinct clustering methods for constructing a hierarchy were explored. The first method involved estimating the conditional means and clustering them using Gaussian Mixture Models. The second method entails measuring the pairwise task similarity between conditional distributions and utilizing a combination of spectral embedding and Gaussian Mixture Models for clustering. 

The aim of this study is to examine the impact of assuming the presence of a hierarchy in time-series data, even when it is not directly defined in the label set. The process of revealing the hierarchical structure is combined with the technique of constructing a hierarchy through hierarchical clustering. The findings reveal that this assumption is advantageous in the some of the datasets in UCR archive \cite{dau2019ucr}. The results suggest that incorporating a hierarchy as a pre-processing step proves beneficial when employing off-the-shelf classifiers for multi-class time series problems. Additionally, there is potential for exploring numerous other hierarchical schemes that can be applied to time series data.

The contributions of this work can be summarized as follows:

\textit{Proposed Hierarchical Divisive Clustering Approach:} The work introduces a novel hierarchical divisive clustering approach with SSFs to improve classification performance in multi-class datasets. This method systematically constructs a hierarchical tree by dividing classes into subsets based on their similarity, optimizing the hierarchical organization without requiring explicit hierarchy information.

\textit{Enhancement of Classification Performance:} The proposed approach demonstrates substantial improvements in classification performance, more notably when using the specialized time series classifier \texttt{rocket}. By effectively leveraging the hierarchical structure, the approach achieves enhanced classification performance compared to FC.

\textit{Automatic Hierarchy Generation:} The method enables automatic hierarchy generation when explicit hierarchical information is not available. By systematically partitioning classes based on their similarity, the approach offers an efficient solution to construct the hierarchical tree representation.

\textit{Balance Factors for Classes and Datapoints:} The work introduces novel concepts of Balance Factor for Classes (BFC) and Balance Factor for Datapoints (BFD) to characterize the balance within the hierarchical tree structure. These factors provide valuable insights into the distribution of classes and datapoints within the hierarchy.

\textit{Insights into Dataset Features:} The study explores the relationship between dataset features and hierarchical clustering performance. It identifies the number of classes and FC score as significant factors for specific classifiers, offering valuable insights into the impact of dataset characteristics on clustering outcomes.

\textit{Efficiency and Efficacy Considerations:} The work acknowledges the stochastic nature of the approach and highlights potential limitations related to efficiency, convergence, solution quality, and sampling bias. The discussion paves the way for future research to explore more efficient optimization techniques and strategies to strike a balance between stochasticity and convergence.

\textit{Potential for Practical Adoption:} The proposed approach is implemented using standard programming libraries and tools, making it relatively easy to implement. The open-source availability of the source code\footnotemark{} and detailed explanations in the study materials enhance the accessibility and usability of the approach for researchers and practitioners interested in exploring hierarchical clustering methods. The results in this paper are reproducible and the code can be easily adapted to work with different classifiers and different datasets.

\footnotetext{\url{https://github.com/alagoz/hc4tsc_hdc_ssf}}

\begin{figure*}[ht]
  \centering
  \includegraphics[width=\textwidth]{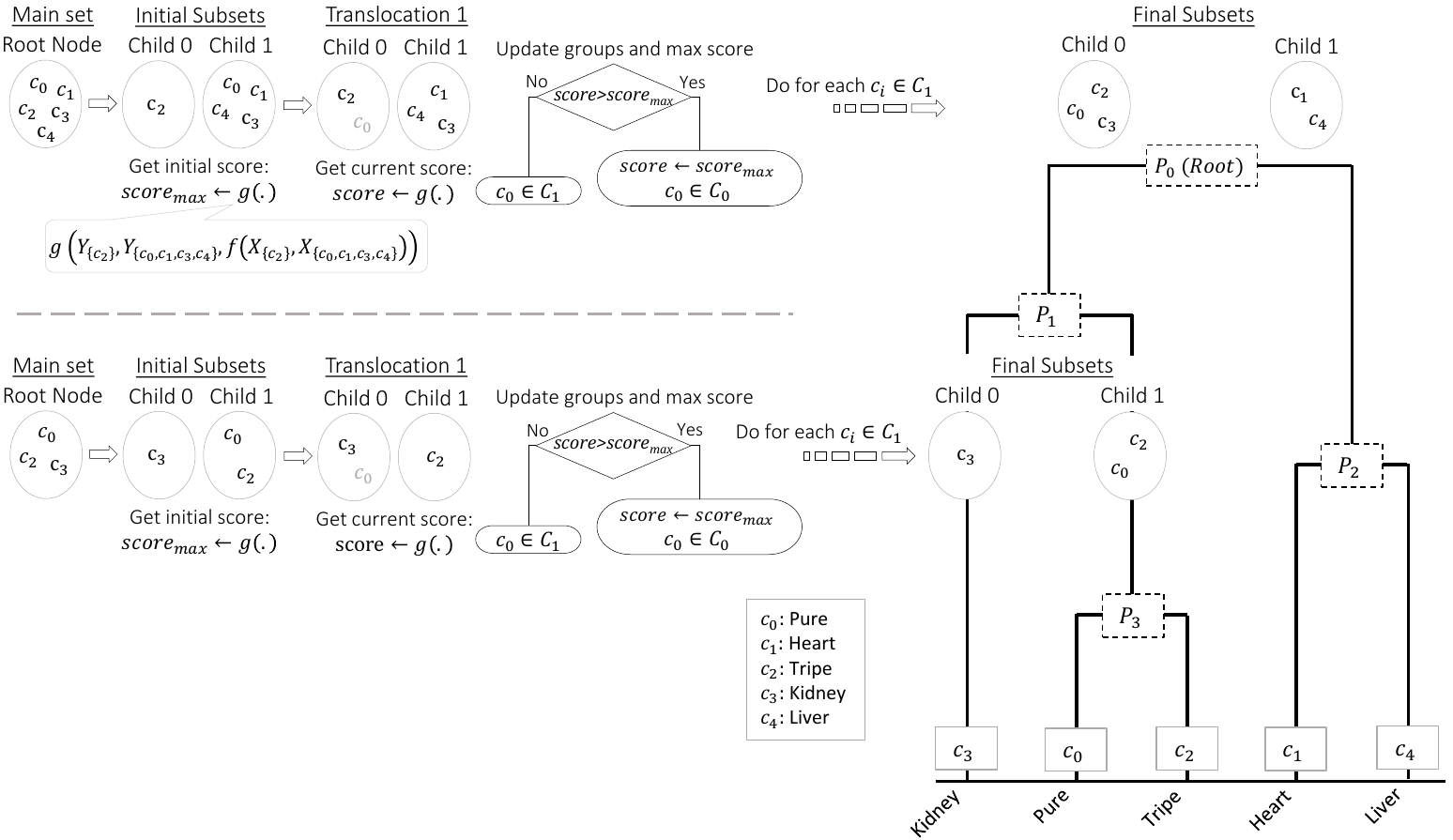}
  \caption{Demonstration of hierarchical divisive clustering with stochastic splitting using the pick-one-then-regroup (\texttt{potr}) algorithm through an example case, the "Beef" dataset from UCR archive. The process involves two iterations: first, the root node is partitioned, and then node 1. Nodes 2 and 3 did not require further partitioning. Since a LCPN approach is used, classifiers are placed at non-singleton cluster nodes only. The nodes with classifiers are sketched with dashed lines.}
  \hypertarget{fig:main}{}
  \label{fig:main}
\end{figure*}

\section{Background}
\label{sec:background}
\subsection{Hierarchical Classification}
The primary area of focus in machine learning research has been on developing models for typical classification problems, where an object is assigned to a single class from a set of non-overlapping classes. This type of classification in the present work is called FC. However, there is a specialized category of tasks where classes are not non-overlapping, but instead organized into a hierarchy. This is known as HC, where objects are associated with a superclass (or parents) and its corresponding subclasses (or children), and the correspondence may be with all or only some of the subclasses. One distinct feature that HC exhibits compared to regular classification is that the classes are structured in a hierarchical manner, meaning that an example belonging to a particular class automatically belongs to all of its superclasses. This is known as  \emph{hierarchical constraint}.




HC is characterized by three attributes \cite{sun2001hierarchical,silla2011survey}:
One critical aspect to consider is the representation of the hierarchical classes, which are depicted as nodes in the graph, and their interrelationships represented as edges in the graph. This attribute can take on either a tree (T) or a Directed Acyclic Graph (DAG) form. In this study, the datasets are structured hierarchically in the form of a tree.

Another aspect to consider is whether a data instance is permitted to have class labels associated with a single path or multiple paths in the class hierarchy. In the case of a single path, only one path of labels is allowed within the hierarchy. Conversely, in the case of multiple paths, the problem involves instances that have more than one path of labels in the hierarchy. In particular, a single example can be associated with multiple classes concurrently. This labeling scenario is commonly referred to as a Hierarchical Multi-Label Classification (HMC) problem in the literature. This study deals with HC only, does not consider the multi-label option.

The depth of classification within the hierarchy is another key aspect to consider. It refers to whether the output of the classifier always corresponds to a leaf node, known as Mandatory Leaf-Node Prediction, or if the predicted class can be positioned at any level within the hierarchy, referred to as Non-Mandatory Leaf Node Prediction. This study focuses specifically on addressing the problem of mandatory leaf node prediction.

There are four approaches of HC \cite{silla2011survey} in terms of how classifiers are deployed. First is called global or big-bang HC approach where all hierarchical classes are assigned to a single classifier. Local classifier per level (LCL) trains a multi-class classifier for each level at hierarchy. Local classifier per node (LCN) approach trains a binary classifier for each hierarchical class that excludes root node. Local classifier per parent node (LCPN) trains a multi-class (a binary one in case of binary tree) classifier for each parent node that includes the root node. This study adopts LCPN approach.

\subsection{Time Series Classification}
Time series classification is a subfield of machine learning that deals with the problem of assigning a label to a sequence of data points over time. In other words, time series classification aims to identify the category or class to which a given time series belongs, based on its temporal behavior. The UCR Time Series Classification Archive is a collection of over 128 time series datasets that have been widely used as benchmarks for evaluating the performance of time series classification algorithms. The datasets cover a wide range of application domains, including biomedicine, finance, and manufacturing, among others.




\section{Approach}
In this study, the terms \emph{category} or \emph{group} are used to refer to a set of classes. Similarly, the terms \emph{sub-class}, \emph{sub-group}, \emph{sub-sets}, and \emph{children} are used interchangeably to denote a subset or subdivision within a category. Likewise, the terms \emph{super-class}, \emph{super-group}, \emph{super-set}, and \emph{parent} are used interchangeably to refer to a higher-level category or set that encompasses the sub-classes or sub-groups. These terms are employed throughout the study to describe the hierarchical relationships and divisions within the class structure.


To provide a foundation for the upcoming sections, several definitions and premises are presented. Consider a dataset denoted by $X=\{x_i,y_i\}_{i=0}^{|X|\text{-}1}$, where $X$ represents the set of data points and $Y$ represents the set of corresponding class labels. Each data point $x_i$ is associated with a class label $y_i$ drawn from a set of $|C|$ unique class labels, denoted as $C$.

In the context of this study, certain terms and notations are introduced to describe the hierarchical structure. A subscript used for a dataset $X$ denotes a subset of the dataset. For example, $X_{c_i}$ refers to the set of data points associated with the class label $c_i$. When a set includes subsets, as is the case for representing parent nodes in this study, the set exhibits a nested structure. In this context, a class in a subset is denoted as $P_{i,j}$, where the subscript after the comma represents the elements within the nested subset.

To differentiate between the set of flat class labels and the set of hierarchical class labels, the notations $C$ and $P$ are used, respectively. Since the hierarchy is represented as a binary tree, the following premises hold:
(i) $P_0$ represents the root node of the tree,
(ii) The number of parent nodes, denoted as $|P|$, is one less than the number of flat labels, denoted as $|C|$,
(iii) The total number of nodes, including both parent and leaf nodes, is $2|C|\text{-}1$,
(iv) Each subset of $P$, denoted as $P_i$, contains exactly two elements, i.e., $|P_i|=2$, where $i$ ranges from 0 to $|P|\text{-}1$. This property arises because each parent node has both a left child and a right child.

In the context of classification and evaluation, the following notations are used:
$X_{tr}$ and $X_{te}$ represent the training and testing subsets of the dataset $X$, respectively.
$X^k$ denotes the kth fold of the dataset in a k-fold division.
For prediction using a flat classifier algorithm, the notation $f(X)$ is used, indicating the prediction made by the flat classifier on the dataset $X$. For prediction using a hierarchical classifier algorithm, the notation $h(P, X)$ is used, where $P$ represents the hierarchy and $X$ is the dataset. This notation denotes the prediction made by the hierarchical classifier algorithm on the dataset $X$ considering the given hierarchy $P$. The evaluation of a classifier is denoted as $g(Y, f(X))$, where $Y$ represents the true class labels and $f(X)$ represents the predicted class labels by the classifier. This notation represents the evaluation of the classifier's performance by comparing the predicted labels with the true labels. 

\begin{algorithm}
\DontPrintSemicolon
\caption{\texttt{fitLcpnTree}: Hierarchical divisive clustering with stochastic splinter}\label{alg:main} \hypertarget{alg:main}{}
\KwData{A dataset with $X$ and $Y$}
\KwResult{A binary tree represented as a set of non-singleton (parent) nodes P}
$C \gets \texttt{extractFlatLabels}(Y)$\;
$P_0 \gets C$\;
\For{$i\,\in\,\{\,0\,...\,|C|\text{-}2\,\}$}{ 
  \If{$|P_i|$ > 2}{
    $P_{i,0},\,P_{i,1} \gets \texttt{splitFun}(X,Y,P_i)$\; 
  }
  set $P_i$ as parent for $P_{i,0}$ and $P_{i,1}$\;
  \If{$|P_{i,0}|>|P_{i,1}| \; \& \; |P_{i,0}|>2$}{
    $P_{i+1} \gets P_{i,0}$\; 
  }
  \ElseIf{$|P_{i,1}|>|P_{i,0}| \; \& \; |P_{i,1}|>2$}{
    $P_{i+1} \gets P_{i,1}$\; 
  }
}
\end{algorithm}

The problem at hand is to generate a hierarchy of classes $P$ from a set of flat classes $C$. The process is outlined in Algorithm \hyperlink{alg:main}{1}. The algorithm begins by assigning the set of flat classes to the root node. Then, it iteratively partitions each parent node into two clusters, following a top-down clustering approach. An example procedure is depicted in Figure \hyperlink{fig:main}{1}. The height of a parent node corresponds to the number of classes present under that node. Parent nodes that already have two elements do not need to be further partitioned. The next parent node to be partitioned is determined based on its height, with the highest one chosen first. This approach ensures the monotonicity of the resulting tree.

The output of the algorithm is a set of parent nodes, each of which has exactly one parent (except for the root node) and two children. As an example, tree in Figure \hyperlink{fig:main}{1} is represented as $P=\{\{\{c_1,c_4\},\{c_0,c_2,c_3\}\},\{\{c_3\},\{c_2,c_0\}\},\{\{c_1\},\{c_4\}\},\{\{c_2\},\{c_0\}\}\}$. The parent-child relationship is implicit in the set $P$. If a subset includes all the elements of another subset, then the former subset is considered the parent of the latter. This relationship is more explicitly represented in the form of a tree object in the source code.\footnotemark[\value{footnote}]

To analyze the resultant hierarchy represented as a binary tree, it is important to characterize the balance of the tree. The term \emph{tree balance} can refer to the balance between the number of left and right descendant nodes \cite{donald1999art} or class labels, or the balance between the number of data instances for the root node. In this study, two metrics are introduced to the literature to assess the tree balance: BFC and BFD which are defined as:
\begin{equation}
BFC(P) = \sum_{\omega \in P}(|\omega_1|\,\text{-}\,|\omega_0|)/\sum_{\omega \in P}(|\omega_1|\text{+}|\omega_0|\text{-}2) \\
\end{equation}
\begin{equation}
BFD(P,X) = \sum_{\omega \in P}(|X_{\omega_1}|\,\text{-}\,|X_{\omega_0}|)/\sum_{\omega \in P}(|X_{\omega_1}|\text{+}|X_{\omega_0}|\text{-}2) 
\end{equation}

The concept of BFC in this study is similar to the balance factor proposed in \cite{donald1999art}, but with some modifications. BFC focuses on the balance between the number of class labels in the left and right branches of the tree throughout all parent nodes not for only the root noade as it was considered in \cite{donald1999art}. Similarly, BFD assesses the balance in terms of the number of data instances. Additionally, contribution of each parent node to the balance in both BFC and BFD is weighted with the number of elements in that node.

By comparing the number of class labels or data instances in the left and right branches of each parent node, BFC and BFD provide a measure of the balance or imbalance within the hierarchical tree structure. The normalization of BFC and BFD ensures that the values are within the range of [-1, 1]. A negative value indicates a left-heavy tree or an imbalance towards the left branch, while a positive value indicates a right-heavy tree or an imbalance towards the right branch. A value of 0 indicates a balanced tree with an equal distribution of class labels or data instances between the left and right branches. It is important to note that for a parent node $P_i$, $P_{i,0}$ and $P_{i,1}$ represent the set of classes under the left and right child nodes, respectively.

After the generation of the hierarchical tree, the HC process is carried out. The training phase of HC is straightforward, where each classifier trains independently based on the assigned classes at each parent node in the tree. This makes the training phase highly efficient, especially when leveraging multiprocessing techniques. This study employs \texttt{joblib.parallel} module in the Joblib library using 'locky' backend for the process-based parallelism. More details can be found in the source code.\footnotemark[\value{footnote}]

However, in the prediction phase, there is interdependence among the classifiers. This is because test instances are passed from ancestors to descendants in the hierarchical tree structure. During the prediction phase, the test instances traverse the tree from the root node to the leaf nodes, following the hierarchical structure. The predictions at each node depend on the predictions made by the classifiers at the parent nodes. This interdependence among classifiers introduces some computational challenges, as the predictions need to be coordinated and propagated through the tree structure.

\subsection{Divisive Clustering via Stochastic Splitting}
The desired hierarchy of classes in the proposed approach is represented as a binary tree, which is a set of non-singleton clusters (parent nodes) denoted as $P$. This hierarchical structure is used in the LPCN approach where classifiers are situated at parent nodes. The top-down, also known as divisive, approach is used for grouping similar classes in the hierarchy.

The rationale behind the divisive approach is that errors or misclassifications at the higher levels of the hierarchy can have a cascading effect on the lower levels. In other words, misclassifications at the top of the hierarchy can propagate down to the bottom, but misclassifications at the bottom do not affect the higher levels. Therefore, by grouping similar classes from the top of the hierarchy and proceeding downwards, the impact of misclassifications can be minimized and the overall performance of the HC can be improved. Overall, the top-down approach in constructing the hierarchy is considered more optimal in terms of minimizing error propagation and capturing meaningful relationships between classes in the classification problem.

One crucial aspect of the proposed approach is the clustering of similar classes. To achieve this, SSFs are introduced. These splitting functions play a key role in the tree generation process and determine how the classes are partitioned at each parent node. The following section is dedicated to describing these SSFs in detail.

\subsubsection{SSFs}
This section introduces a family of splitting functions that are used to divide a given set of classes into two subsets, with the aim of grouping similar classes together. The notion of similarity in this context refers to the difficulty of discriminating between those classes using the classifier being employed.

The objective is to maximize the classification score between the formed groups. This can be formally defined as:

\begin{equation}
\argmax_{i} g\,(\{Y_{c_i},Y_{c_j}\},f(\{X_{c_i},X_{c_j}\})),\; \forall_{i}, i\neq j 
\end{equation}

The aim is to find a partition of the classes that maximizes this objective function. Since there are $2^{|C|\text{-}1}\text{-}1$ possibilities to form two non-empty subsets from a set of classes with $|C|$ members, exhaustively trying all possibilities becomes computationally prohibitive for larger numbers of classes because the computational cost grows exponentially. Therefore, more efficient approaches are needed. The following algorithms discuss alternative methods that can achieve the desired splitting while reducing computational complexity. These algorithms aim to strike a balance between computational efficiency and the quality of the obtained partition.

\paragraph{Pick-one-then-regroup:} The algorithm, described in Algorithm \hyperlink{fig:main}{2}, starts by randomly selecting an element from the given set and placing it in one cluster, while the remaining elements are placed in the other cluster. The initial classification performance of this partition is recorded. Next, each remaining element is individually experimented with, by translocating them. The algorithm monitors the change in classification performance for each experiment. If there is an improvement in the current performance, the experimented member stays in the present cluster, and the maximum score is updated. This iterative process continues for all remaining elements, evaluating the potential improvements in the classification performance by transferring them to the other cluster. Figure \hyperlink{fig:main}{1} provides an example to illustrate the pick-one-then-regroup approach.

\begin{algorithm}
\DontPrintSemicolon
\caption{\texttt{potr}: Pick-one-then-regroup}\label{alg:potr}
\hypertarget{alg:potr}{}
\KwData{A dataset with $X$ and $Y$ and set of classes $C$ to be partitioned}
\KwResult{Siblings $C_0$ and $C_1$ partitioned from $C$ }
Let $j$ be a random variable $\in [0,|C|\text{-}1] \cap \mathbb{Z}$\;
$C_0 \gets \{C_j\}$\;
$C_1 \gets \{C_i\}_{i\neq j}$\;
$score_{max} \gets g(\{Y_{C_0},Y_{C_1}\},f(\{X_{C_0},X_{C_1}\}))$\;
\For{$c \in \, C_1$}{
  $C_0' \gets C_0 \cup c$\;
  $C_1' \gets C_1 \setminus c$\;
  $score \gets g(\{Y_{C_0'},Y_{C_1'}\},f(\{X_{C_0'},X_{C_1'}\}))$\;
  $C_0, C_1, score_{max} \gets$ \texttt{updateScoreAndGroups}()\;
}
\end{algorithm}

\paragraph{Split-randomly-then-regroup:} The algorithm described in Algorithm \hyperlink{fig:main}{3} begins by shuffling the given set of classes and splitting it from a random point. The initial classification performance of this partition is recorded. Next, each element is individually translocated, with the condition that there will be no empty set after the translocation. The algorithm then monitors the change in classification performance resulting from each translocation. If there is an improvement in the classification performance, the element remains in the set where it was transferred. The algorithm continues this process for each element in the set.
\begin{algorithm}
\DontPrintSemicolon
\caption{\texttt{srtr}: Split-randomly-then-regroup}\label{alg:srtr}
\hypertarget{alg:srtr}{}
\KwData{A dataset with $X$ and $Y$ and set of classes $C$ to be partitioned}
\KwResult{Siblings $C_0$ and $C_1$ partitioned from $C$ }
$C \gets$ Shuffle($C$)\;
Let $j$ be a random variable $\in [1,|C|\text{-}1] \cap \mathbb{Z}$\;
$C_0 \gets \{C_i\}_{i<j}$\;
$C_1 \gets \{C_i\}_{i\geq j}$\;
$score_{max} \gets g(\{Y_{C_0},Y_{C_1}\},f(\{X_{C_0},X_{C_1}\}))$\;
\For{$c \in \, C$}{
  {\If{$c \in C_0\; \&\; |C_0|>1$}{
    $C_0' \gets C_0 \setminus c$\;
    $C_1' \gets C_1 \cup c$\;
    }
  }
  {\If{$c \in C_1\; \&\; |C_1|>1$}{
    $C_0' \gets C_0 \cup c$\;
    $C_1' \gets C_1 \setminus c$\;
    }
  }
  $score \gets g(\{Y_{C_0'},Y_{C_1'}\},f(\{X_{C_0'},X_{C_1'}\}))$\;
  $C_0, C_1, score_{max} \gets$ \texttt{updateScoreAndGroups}()\;
}
\end{algorithm}

Overall, both \texttt{potr} and \texttt{srtr} approaches provide a systematic way to find an optimized partition of the classes by iteratively evaluating the impact of translocating elements between clusters and updating the classification performance accordingly. The \texttt{potr} approach reduces the number of possibilities to $|C|\text{-}1$, while the \texttt{srtr} approach reduces it to $|C|$. Both approaches significantly reduce the computational cost compared to exhaustively considering all possible partitions, resulting in linear computational complexity.

By considering the performance improvement achieved by translocating elements, both approaches efficiently find a partition that maximizes the classification performance. These approaches offer efficient alternatives to handle the exponential number of possibilities and effectively identify an optimal partition of the classes.

\paragraph{Leave-salient-one-out:} This approach specifically leaves one member out rather than translocating the members between clusters. As described in Algorithm \hyperlink{fig:main}{4}, it starts with shuffling the given set and iteratively leaves each member out. The member that results in the maximum classification performance when it is left out is identified. The \texttt{lsoo} approach provides an efficient way to find an optimized partition of the classes by iteratively evaluating the impact of leaving each member out and selecting the optimal configuration. The leave-one-out approach reduces the number of possibilities to $|C|$, resulting in a linear computational complexity.
\begin{algorithm}
\DontPrintSemicolon
\caption{\texttt{lsoo}: Leave-salient-one-out}\label{alg:lsoo}
\hypertarget{alg:lsoo}{}
\KwData{A dataset with $X$ and $Y$ and set of classes $C$ to be partitioned}
\KwResult{Siblings $C_0$ and $C_1$ partitioned from $C$ }
$C \gets$ shuffle($C$)\;
$score_{max} \gets 0$\;
\For{$c \in \, C$}{
  $C_0' \gets {c}$\;
  $C_1' \gets C \setminus c$\;
  $score \gets g(\{Y_{C_0'},Y_{C_1'}\},f(\{X_{C_0'},X_{C_1'}\}))$\;
  $C_0, C_1, score_{max} \gets$ \texttt{updateScoreAndGroups}()\;
}
\end{algorithm}

A maximum score and the child groups update procedure (see Algorithm \hyperlink{alg:update}{5}) is used in all split functions. If the current score is better than the score which is maximum so far, then groups and the maximum score is updated. Also, a stopping criterion is implemented whenever a performance of $100\%$ achieved. In this case, there is no further need to proceed to seek the best split, and the algorithm terminates. This then reduces the number of possibilities to even less than $|C|$.
\begin{algorithm}
\DontPrintSemicolon
\caption{\texttt{updateScoreAndGroups}: Routine to update score and groups}\label{alg:update}
\hypertarget{alg:update}{}
\KwData{Maximum score $score_{max}$, current score $score$, and siblings $C_0'$ and $C_1'$}
\KwResult{Maximum score $score_{max}$ and siblings $C_0$ and $C_1$}
{\If{$score > score_{max}$}{
  $score_{max} \gets score$\;
  $C_0 \gets C_0'$\;
  $C_1 \gets C_1'$\;
  {\If{$score == 100$}{
    break;}
  }}
}
\end{algorithm}

\subsection{Evaluation, Hyperparameters, and Cross Validation}
In this study, two different classifiers were utilized: \texttt{rocket} \cite{dempster2020rocket} and \texttt{svm} with a linear kernel \cite{cortes1995support}. These classifiers were chosen for their efficiency and suitability for time series classification tasks.

For the \texttt{rocket} classifier, the number of kernels was set to 512. This choice allows for a balance between computational efficiency and capturing relevant features from the time series data. 

To ensure compatibility and reproducibility, the random number generator seed was set to a default value, which in this case is zero. Setting the seed value helps in obtaining consistent results across different runs of the experiment.

The classification performance was evaluated using the f1 score, which is a commonly used metric for binary classification tasks. In the case of multi-class classification, the f1 macro score was utilized. The f1 macro score calculates the f1 score for each class and then takes the average, giving equal weight to each class. By employing the f1 score and f1 macro score as evaluation metrics, the study assesses the classification performance of the chosen classifiers in a comprehensive and robust manner.

The proposed hierarchy generation model makes use of stochastic splitting algorithms, which require executing the tree generation procedure multiple times in order to find a suboptimal tree. The number of iterations is manually specified and experimented with different values to find the best tree, which corresponds to the model's hyperparameter. The goal is to maximize the performance of the model on unseen data, represented by the test set. To achieve this, a nested cross-validation (CV) approach, also known as double CV \cite{stone1974cross}, is employed.

The nested CV consists of an outer and an inner CV loop (see Algorithm \hyperlink{alg:nCV}{6}). The outer loop is responsible for partitioning the dataset into training and testing subsets. In this study, a 5-fold cross-validation is applied, where in each iteration of the outer loop, one fold is held out as the test set, and the remaining 4 folds are used as the training set.
\begin{algorithm}
\DontPrintSemicolon
\caption{Nested CV procedure}\label{alg:nested}
\hypertarget{alg:nCV}{}
\KwData{A dataset with $X$ and $Y$}
\KwResult{Nested CV score $score\_nCV$ and set of selected trees $T$}
$scores\_out \gets \{\}$\;
$T \gets \{\}$\;
\For{$ko \in \, \{1,...,N_{te}\}$}{
  $X^{ko}, Y^{ko} \gets \texttt{splitData}(X,Y,$ shuffle=$0)$\;
  $scores\_out_{ko} \gets 0$\;  
  \For{$i \in \, \{1,...,N_{iter}\}$}{
    $X^i, Y^i \gets \texttt{splitData}(X^{ko}_{te}, Y^{ko}_{te},$ shuffle=$1)$\;
    $P \gets \texttt{fitLcpnTree}(X^i, Y^i)$\;
    $\texttt{checkDuplicatesAndLimit}()$\;
    $scores\_in \gets \{\}$\;
    \For{$ki \in \, \{1,...,N_{va}\}$}{
      $X^{ki}, Y^{ki} \gets \texttt{splitData}(X^{ko}_{te}, Y^{ko}_{te},$ shuffle=$0)$\;
      $scores\_in_{ki} \gets g(Y^{ki}_{te},h(P,X^{ki}_{te}))$\;      
    }
    {\If{$\mathrm{mean}(scores\_in) > scores\_out_{ko}$}{
    $scores\_out_{ko} \gets$ mean($scores\_in$)\;
    $T_{ko} \gets P$\;
    }
    }
  }
}
$score\_nCV=$ mean($scores\_out$)\;
\end{algorithm}

Before the inner loop, there is an additional loop to iterate over the specified number of iterations for the model. This results in two levels of nested loops in total. The loop in the middle and the innermost loop are specifically designed to find the best tree.

In each iteration of the loop in the middle, a new tree is generated using the training set from the outer loop. The training set is shuffled and then divided into 4 folds, with one fold used as the validation set and the remaining 3 folds used for training.

For each generated tree, the innermost loop calculates an inner CV score. This is done by performing 4 iterations, where each iteration uses one fold as the validation set and the remaining 3 folds as the training set. The mean score of the inner CV represents the algorithm's performance on different subsets of the training data. The goal is to select the optimal tree that maximizes this mean score. By evaluating the performance of the algorithm on multiple subsets of the training data, the mean score provides an estimate of how well the algorithm generalizes to unseen data. The tree with the highest mean score is considered the optimal tree, as it demonstrates the best performance on the validation sets within the inner CV loop.

By employing this nested CV approach, the proposed model can evaluate and compare the performance of different trees while accounting for variations in the training and validation subsets. This helps to mitigate potential overfitting and provides a more reliable estimate of the model's performance on unseen data.

To assess the generalization performance of the model, in addition to the nested CV scheme, a separate scheme known as flat CV, as described in the work by Wainer (2021) \cite{wainer2021nested}, is employed. The flat CV scheme involves dividing the dataset into multiple folds using a single k-fold CV step, without employing a nested loop. Each flat CV fold iteration is dedicated to both selecting the optimal hyperparameters and evaluating the average performance across all folds. It is important to note that this estimation may be biased, but previous research has shown that it can still be used to determine the best hyperparameters, which aligns with the results obtained from nested CV in certain cases \cite{wainer2021nested}. However, in this particular work, the flat CV scheme is not utilized to assess the generalization performance of the proposed model. Rather, its purpose is to compare the performance of the flat CV approach with that of the nested CV approach. This comparison provides insights into the relative performance of the proposed model when evaluated using different CV schemes.

Flat CV used in this study has two loops (see Algorithm \hyperlink{alg:fCV}{7}). Similar to nested CV, the outer loop is responsible for partitioning the dataset into training and testing subsets. The inner loop, on the other hand, is dedicated to tree generation and iterates over a specified number of iterations.
\begin{algorithm}
\DontPrintSemicolon
\caption{Flat CV procedure}\label{alg:fCV}
\hypertarget{alg:fCV}{}
\KwData{A dataset with $X$ and $Y$}
\KwResult{Flat CV score $score\_fCV$ and selected trees $P$}
$scores\_out \gets \{\}$\;
$T \gets \{\}$\;
\For{$ko \in \, \{1,...,N_{te}\}$}{
  $X^{ko}, Y^{ko} \gets \texttt{splitData}(X,Y,$shuffle=$0)$\;
  $scores\_out_{ko} \gets 0$\;  
  \For{$i \in \, \{1,...,N_{iter}\}$}{
    $X^i, Y^i \gets \texttt{splitData}(X^{ko}_{te}, Y^{ko}_{te},$shuffle=$1)$\;
    $P \gets \texttt{fitLcpnTree}(X^i, Y^i)$\;
    $\texttt{checkDuplicatesAndLimit}()$\;
    $score \gets g(Y^{ko}_{te},h(P,X^{ko}_{te}))$\;    
    {\If{$score > scores\_out_{ko}$}{
    $scores\_out_{ko} \gets score$\;
    $T_{ko} \gets P$\;    
    }
    }
  }
}
$score\_fCV=$ mean($scores\_out$)\;
\end{algorithm}

In the inner loop of flat CV, the training set is shuffled and divided into 4 folds. One fold is used as the validation set, while the remaining 3 folds are utilized for training. It is important to note that tree generation is performed using both the training and validation sets, without including the holdout data. The evaluation of the generated trees, however, is conducted using the test data. This is where the bias in the flat CV approach arises. While the tree generation process itself is unbiased, the evaluation of the selected tree is influenced by the use of the test data. This bias stems from the fact that the selected tree is evaluated on the same data that was used for testing, potentially leading to an overly optimistic estimation of the model's performance. Overall, the flat CV approach used in this study introduces bias in the evaluation of the selected tree, but not in the tree generation process itself.

To optimize computational efficiency during the process of generating multiple trees and finding the optimal tree, two critical aspects are considered: managing duplicate trees and determining when to stop the iteration based on the total number of distinct trees processed. The \texttt{checkDuplicatesAndLimit} function serves the purpose of addressing these considerations. It performs two essential tasks:

\paragraph{Handling Duplicate Trees:}To handle duplicate trees during the iteration process, a notion of "tree similarity" is utilized specific to the problem at hand. It is important to note that tree similarity is not the same as tree equality, where the positions of all parent and leaf nodes need to be identical. Instead, tree similarity focuses on whether the parent nodes contain the same classes, disregarding the order of the classes within the subgroups as well as super-groups. For example, consider two trees: $P^1=\{\{\{c_0,c_1\},\{c_2,c_3\}\},\{\{c_0\},\{c_1\}\},\{\{c_2\},\{c_3\}\}\}$ and $P^2=\{\{\{c_3,c_2\},\{c_1,c_0\}\},\{\{c_3\},\{c_2\}\},\{\{c_1\},\{c_0\}\}\}$. Despite having different class orders within the subgroups as well as super-groups, these trees are considered similar. To detect tree similarity, a pre-order traversal of the trees is performed. The corresponding parent nodes of the trees are then compared to determine whether they contain the same classes, irrespective of their order. This comparison allows for the identification of similar trees and helps in managing duplicates during the iteration process. For a more detailed understanding of the implementation and the exact code logic, referring to the source code is recommended.\footnotemark[\value{footnote}] It provides further insights into how tree similarity is assessed and utilized to handle duplicate trees efficiently.

\paragraph{Iteration Termination:}The function includes a mechanism to determine when to stop the iteration based on the total number of distinct trees processed reaches a certain threshold. This enables the termination of the iteration process once the total number of distinct trees has been reached, effectively saving computational resources. This threshold is determined based on the number of classes in the problem and can be recursively evaluated as:
\begin{equation}
T_{|C|} = {|C| \choose |C|\text{-}1}T_{|C|\text{-}1} + {|C| \choose |C|\text{-}2}T_{|C|\text{-}2} + ... + \alpha \, {|C| \choose \ceil*{|C|/2}}T_{\ceil*{|C|/2}} 
\end{equation}
where $\alpha$ is 1 if $|C|$ is odd and 1/2 otherwise. Due to recursive nature of the estimation, computational cost grows unbounded with increasing number of classes. To mitigate this issue and save computational resources, a lookup table is utilized. This lookup table helps to store and retrieve previously computed results, eliminating the need for redundant computations. For further details on the implementation of the program, more information can be found in the source code.\footnotemark[\value{footnote}]

In summary, the proposed approach involves using 4-fold-within-5-fold nested CV and 5-fold flat CV to compare with the nested CV and assess an evaluation of model performance. In both CV procedures, the best tree is selected for each training set from the most outer loop, which consists of 5 training sets in this study. Each of these CV procedures includes an additional loop for hyperparameter search, a common practice in hyperparameter tuning. It is worth noting that the hyperparameter selection loop performs data partitioning by shuffling the available data. However, apart from this specific partitioning performed in the hyperparameter selection loop, all other partitions across all datasets remain fixed. This ensures compatibility between the flat and HC schemes and enables an accurate comparison between them.

A stratified split is utilized in all partitions of the dataset. This means that the splitting process ensures that each partition maintains the same class distribution as the original dataset. 

\subsection{Computational Complexity Analysis} \label{sec:cost} \hypertarget{sec:cost}{}
Computational cost can be separately considered for pre-processing, training, and prediction phases. Determining the time complexity of the problem is not straightforward since there are several factors that can not be controlled directly. Some of those can be listed as (i) split functions are non-deterministic and early stop conditions may occur, (ii) data-points allocation varies with tree structure, (iii) the dataset may exhibit class imbalance, indicating that there is an uneven distribution of data points among the different classes, (iv) classifier performance during testing phase affects the pathway for the testing data instances. On the other hand, it is possible to exploit the preprocessing and the training phases to utilize process-based parallelism.


The following notations will be helpful in the clarity and comprehensibility of the forthcoming analysis. Let $M$ be time-points length of data instances and let $\phi_{tr}$ and $\phi_{te}$ represent time complexity of a classifier per each data instance with a single time point (i.e. length of 1) during training and testing phases, respectively. Let $N_{iter}$ denote the number of iterations specified for the stochastic tree generation procedure.

Tree generation is considered as the preprocessing phase, where an iterative splitting operation is carried out using a top-down approach. In the first iteration, all data instances from the training set are utilized. In the subsequent iterations, the total number of datapoints to be processed can vary depending on the balance of datapoint allocation resulting from the previous iteration, specifically from the parent node. In particular terms, the more datapoints are grouped in an imbalanced way, the more datapoints are processed by the classifier in question. Therefore, a minimum number of datapoints to be processed can be approximated assuming each parent maintains a perfect balance of cluster allocation, on the contrary, a maximum number of datapoints to be processed can be approximated assuming maximum imbalance of cluster allocation at each parent node where a single datapoint falls into one clusters while the rest belong to the other:
\begin{equation}
  \approx \begin{cases}
    2|X_{tr}||C|, & \text{if $BFC(P)=BFD(P,X_{tr})=0$}.\\
    |X_{tr}||C|^2/2, & \text{if $BFC(P)=BFC(P,X_{tr})\in\{\text{-}1,1\}$}.
  \end{cases}
\end{equation}

Since the training set splits further into training and validation sets in the preprocessing phase, let $|X_{trval}|$ and $|X_{val}|$ denote the number of data instances in the training used for validation and validation sets, respectively. Hence, for the preprocessing phase, time complexity is expected to vary between $\mathcal{O}\big(N_{iter}|C|M(|X_{trval}|\phi_{tr}+|X_{val}|\phi_{te})\big)$ and $\mathcal{O}\big(N_{iter}|C|^2M(|X_{trval}|\phi_{tr}+|X_{val}|\phi_{te})\big)$. Tree structure has similar effect on the number of datapoints to be processed in the training phase, then the time complexity in the training phase is expected to vary between the values of $\mathcal{O}(|X_{tr}||C|M\phi_{tr})$ and $\mathcal{O}(|X_{tr}||C|^2M\phi_{tr})$, as determined by the aforementioned derivations. It is worth noting that the actual execution time is expected to decrease due to the highly parallelizable nature of hierarchical training. 

As for the testing phase, complexity during prediction of each data instance depends on tree depth of that instance, specifically how many levels it traverse the tree down until it reaches a leaf node. By considering all test instances, an average tree depth per flat class label can be estimated assuming there is no class imbalance. A shortest path average can be approximated assuming maximum balance of class labels at parent nodes (i.e. $BFC(P)=0$) while a longest path average can be approximated assuming maximum imbalance of class labels at parent nodes (i.e. $BFC(P)=\{\text{-}1,1\}$):
\begin{equation}
  \approx \begin{cases}
  log_{2}|C|, & \text{if $BFC(P)=BFD(P,X_{te})=0$}.\\
  |C|/2, & \text{if $BFC(P)\in\{\text{-}1,1\},BFD(P,X_{te})=0$}.    
  \end{cases}
\end{equation}

The time complexity of the testing phase is then expected to vary between $\mathcal{O}(|X_{te}|Mlog_{2}|C|\phi_{te})$ and $\mathcal{O}(|X_{te}|M|C|\phi_{te})$. 

Overall, the complexity of HC is expected to vary between $\mathcal{O}\big(M(|C|(N_{iter}(|X_{trval}|\phi_{tr}+|X_{val}|\phi_{te})+|X_{tr}|\phi_{tr})+log_{2}|C||X_{te}|\phi_{te})\big)$ and $\mathcal{O}\big(M|C|(|C|(N_{iter}(|X_{trval}|\phi_{tr}+|X_{val}|\phi_{te})+|X_{tr}|\phi_{tr})+|X_{te}|\phi_{te})\big)$. These can be compared to the FC cost $\mathcal{O}(M(|X_{tr}|\phi_{tr}+|X_{te}|\phi_{te}))$. The overall expected cost is expected to increase in a respective order with the trees generated using split functions \texttt{srtr}, \texttt{potr}, and \texttt{lsoo}. Specifically, the algorithm \texttt{srtr} is expected to be the most efficient among them.

\section{Experiments on Time Series Data}
The efficacy of the proposed approach is tested on a selection of datasets from the UCR archive \cite{dau2019ucr}. The UCR archive contains a collection of 128 univariate time series datasets. For this study, only multi-class cases with more than two classes are considered. Additionally, datasets with classification accuracy greater than $99.5\%$ for both \texttt{svm} and \texttt{rocket} classifiers were excluded, as there is little room for improvement in such cases.

After applying these criteria, a total of 46 datasets out of the original 128 were selected for evaluation. The selected datasets were downloaded and handled using the \texttt{sktime} library \cite{loning2019sktime}, which provides tools for handling and analyzing time series data. These datasets were used to assess the performance of the proposed approach for HC.

Figure \hyperlink{fig:niter}{2} displays the number of improvements achieved using different classifiers (\texttt{svm} or \texttt{rocket}) and SSFs (\texttt{potr}, \texttt{srtr}, or \texttt{lsoo}) during hierarchical divisive clustering, along with nested or flat CV evaluation schemes. The results were compared for 3, 5, 10, and 50 iterations for both \texttt{rocket} and \texttt{svm}. It can be seen that the number of improvements achieved declines as the number of iterations exceeds 20 for both \texttt{rocket} and \texttt{svm}. For the coming analyses, the focus was on 10 iterations since it seems as a reasonable choice, and the results were included in the Appendix, presented in Table \hyperlink{tab:svm}{2} for \texttt{svm} and Table \hyperlink{tab:rocket}{3} for \texttt{rocket}. These tables display the FC and HC scores obtained using each splitting algorithm, with the improvement state highlighted using bold font weight.
\begin{figure}[ht]
  \centering
  \includegraphics[width=0.5\textwidth]{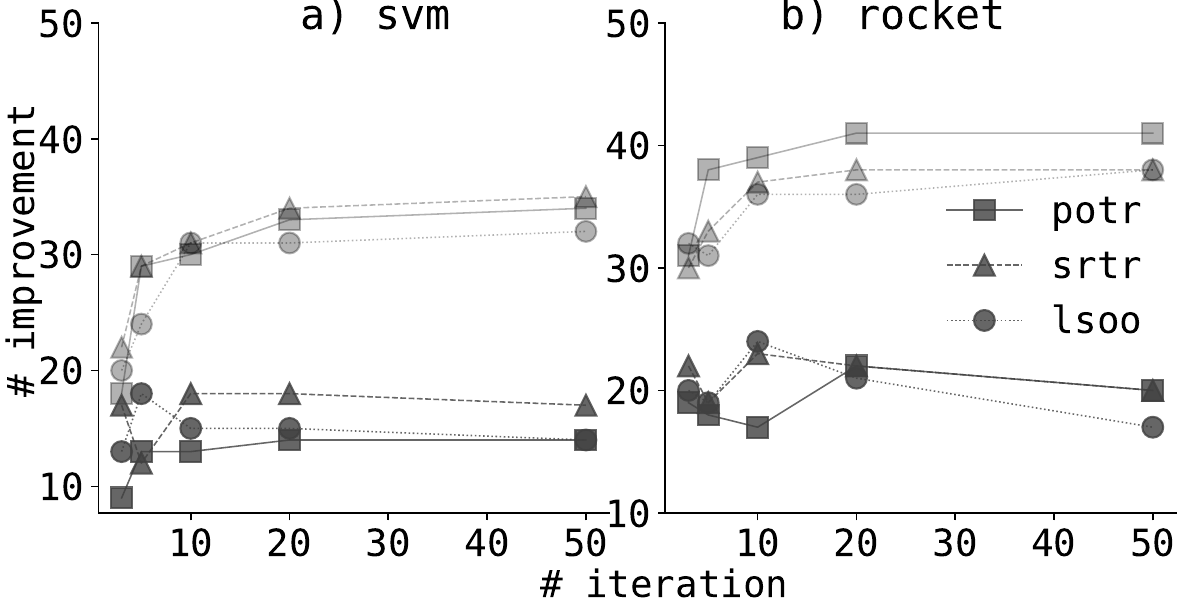}
  \caption{Comparison of the number of iterations versus the number of improved results for \texttt{svm} (a) and \texttt{rocket} (b) when using different splitting functions, such as \texttt{potr}, \texttt{srtr}, or \texttt{lsoo}, and different evaluation schemes, such as nested CV or flat CV. The results obtained with flat CV are shown with lighter marker colors.}
  \hypertarget{fig:niter}{}
  \label{fig:niter}
\end{figure}

In the initial observations, a comparison between the classifiers showed that \texttt{rocket} outperforms \texttt{svm} in terms of the number of improved datasets. As anticipated, the flat CV scheme resulted in more improvements compared to the nested CV scheme. Approximately, \texttt{rocket} achieved improvements in around $50\%$ of the datasets using the nested CV scheme, while \texttt{svm} achieved improvements in around $30\%$. With the flat CV scheme, these numbers increased to approximately $85\%$ and $70\%$ for \texttt{rocket} and \texttt{svm}, respectively. Regarding the splitting functions, all of them yielded overall comparable results.

Regarding the flat CV, the number of improvements generally increases with an increasing number of iterations, but the rate of increase decreases for larger iteration values. This trend suggests that a saturation point might be reached for even higher iterations. On the other hand, in nested CV, the number of improvements fluctuates for smaller iteration values such as 3, 5, and 10. For example, in the case of \texttt{rocket}, the number of improvements is higher when the number of iterations is 3 compared to 5. For \texttt{svm}, a consistent decline in the number of improvements is observed when the number of iterations increases from 20 to 50. This indicates that using a number of iterations greater than 20 is not advisable. For the forthcoming analyses, a feasible choice would be to consider 10 iterations ($N_{iter}=10$) for both classifiers. This range can be narrowed down to 3 to 10 iterations for \texttt{svm}. Hence, 10 iterations will be adopted for the subsequent analyses.

The further analyses involve exploring the relationship between dataset features and the improvement obtained using HC over FC. The considered dataset features include the number of classes and FC score, along with the tree balance metrics BFC and BFD introduced in this study. The outcome of interest can be either the number of improvements or the difference in classification performance, specifically f1 score evaluated by the function $g(.)$, between HC and FC, denoted as $\Delta g$. In other words, the dependent variable can be treated as either a categorical variable (indicating whether there is an improvement or not) or a continuous interval variable. Both types of outputs are considered in this study. 

To assess the association between dataset features and the outcome, statistical tests are conducted. For the continuous interval variable outcome $\Delta g$, Pearson correlation analysis is performed since the independent variables are also continuous interval variables. To obtain a second opinion, multiple linear regression analysis is also implemented. When the outcome is treated as categorical (indicating whether there is an improvement or not), multiple logistic regression analysis is used. The dataset includes 230 samples for the statistical analysis, obtained from the scores during the testing phase of the nested CV scheme for 46 datasets with 5 observations each. The statistical significance threshold is set at $p<0.001$, meaning that any p-value less than 0.001 is considered statistically significant in the analyses.

Table \hyperlink{tab:pcor}{1} presents the results of Pearson correlation tests, including correlation coefficients and corresponding p-values. For the \texttt{svm} classifier, the number of classes is found to be statistically significant in all cases, showing a negative correlation with $\Delta g$. Additionally, the FC score is also statistically significant in all cases for the \texttt{rocket} classifier, demonstrating a negative correlation with $\Delta g$.
\begin{table*}[t!]
 \caption{ Results of the Pearson correlation tests assessing the correlation coefficient and p-values for independence between dataset features and the performance difference between HC and FC (FC), denoted as $\Delta g$. The tests were conducted with $N_{iter}=10$ iterations, considering the individual testing scores of the nested CV scheme. Note that  the test is not applicable to \texttt{lsoo} as the BFC metric always results in a value of 1.)}\label{tab:pcor}
 \hypertarget{tab:pcor}{}
  \centering
  \renewcommand{\arraystretch}{0.5}
  \begin{tabular}{lcccccccccccc}
    \toprule
     & \multicolumn{6}{c}{\texttt{svm}} & \multicolumn{6}{c}{\texttt{rocket}} \\
    \cmidrule(r){2-7} \cmidrule(r){8-13}\\
    & \multicolumn{2}{c}{\texttt{potr}} &\multicolumn{2}{c}{\texttt{srtr}} &\multicolumn{2}{c}{\texttt{lsoo}} &\multicolumn{2}{c}{\texttt{potr}}&\multicolumn{2}{c}{\texttt{srtr}}&\multicolumn{2}{c}{\texttt{lsoo}}\\  
    \cmidrule(r){2-3} \cmidrule(r){4-5} \cmidrule(r){6-7} \cmidrule(r){8-9} \cmidrule(r){10-11} \cmidrule(r){12-13}\\
    Features&$r$&$p$&$r$&$p$&$r$&$p$&$r$&$p$&$r$&$p$&$r$&$p$\\
    \cmidrule(r){1-1} \cmidrule(r){2-2} \cmidrule(r){3-3} \cmidrule(r){4-4} \cmidrule(r){5-5} \cmidrule(r){6-6} \cmidrule(r){7-7} \cmidrule(r){8-8} \cmidrule(r){9-9} \cmidrule(r){10-10} \cmidrule(r){11-11} \cmidrule(r){12-12} \cmidrule(r){13-13} \\
\#class&\textbf{-0.605}&\textbf{0.000}&\textbf{-0.568}&\textbf{0.000}&\textbf{-0.623}&\textbf{0.000}&-0.054&0.417&-0.102&0.124&0.026&0.699\\
FCscore&-0.032&0.625&-0.026&0.696&-0.009&0.888&\textbf{-0.272}&\textbf{0.000}&\textbf{-0.340}&\textbf{0.000}&\textbf{-0.307}&\textbf{0.000}\\
BFD&-0.073&0.272&\textbf{0.209}&\textbf{0.001}&\textbf{-0.510}&\textbf{0.000}&-0.081&0.224&-0.040&0.544&-0.064&0.337\\
BFC&\textbf{0.309}&\textbf{0.000}&\textbf{0.270}&\textbf{0.000}&$-$&$-$&-0.139&0.035&-0.034&0.604&$-$&$-$\\
    \bottomrule
  \end{tabular}
  \label{tab:table}
\end{table*}

For the \texttt{svm} classifier, the BFD metric is positively correlated with $\Delta g$ when using \texttt{srtr}, while it is negatively correlated when using \texttt{lsoo}. On the other hand, for the \texttt{svm} classifier, the BFC metric is positively correlated with $\Delta g$ when using \texttt{potr} and \texttt{srtr} as the SSFs. 

The multiple linear regression analysis results presented in Appendix Table \hyperlink{tab:linreg}{4} confirm the findings from the Pearson correlation analysis regarding the effect of the number of classes on \texttt{svm} and the FC score on \texttt{rocket}. However, for \texttt{svm}, BFC was found to be significant when using only \texttt{potr}, whereas BFD was found to be significant when using only \texttt{lsoo}.

The multiple logistic regression results, which were conducted when the output was considered as categorical, are presented in Appendix Table \hyperlink{tab:linreg}{5}. These results confirm the findings from the Pearson correlation analysis regarding the effect of the number of classes on \texttt{svm} and the FC score on \texttt{rocket}. However, for \texttt{svm}, no significant relationship was found for BFD and BFC. Additionally, in contrast to the Pearson correlation results, the FC score was found to be significant when using \texttt{potr} and \texttt{srtr} as the SSFs.  

Figure \hyperlink{fig:nclass}{3} provides a graphical representation of the comparison between the number of classes and the improvement obtained using HC over FC using a grouped bar chart. For \texttt{svm}, it is evident that the number of improvements decreases as the number of classes increases, which aligns with the negative correlation identified through the Pearson correlation test. Conversely, for \texttt{rocket}, no significant change in the number of improvements is observed with varying the number of classes, consistent with the finding of no significant relation from the Pearson correlation test.

\begin{figure}[ht]
  \centering
  \includegraphics[width=0.5\textwidth]{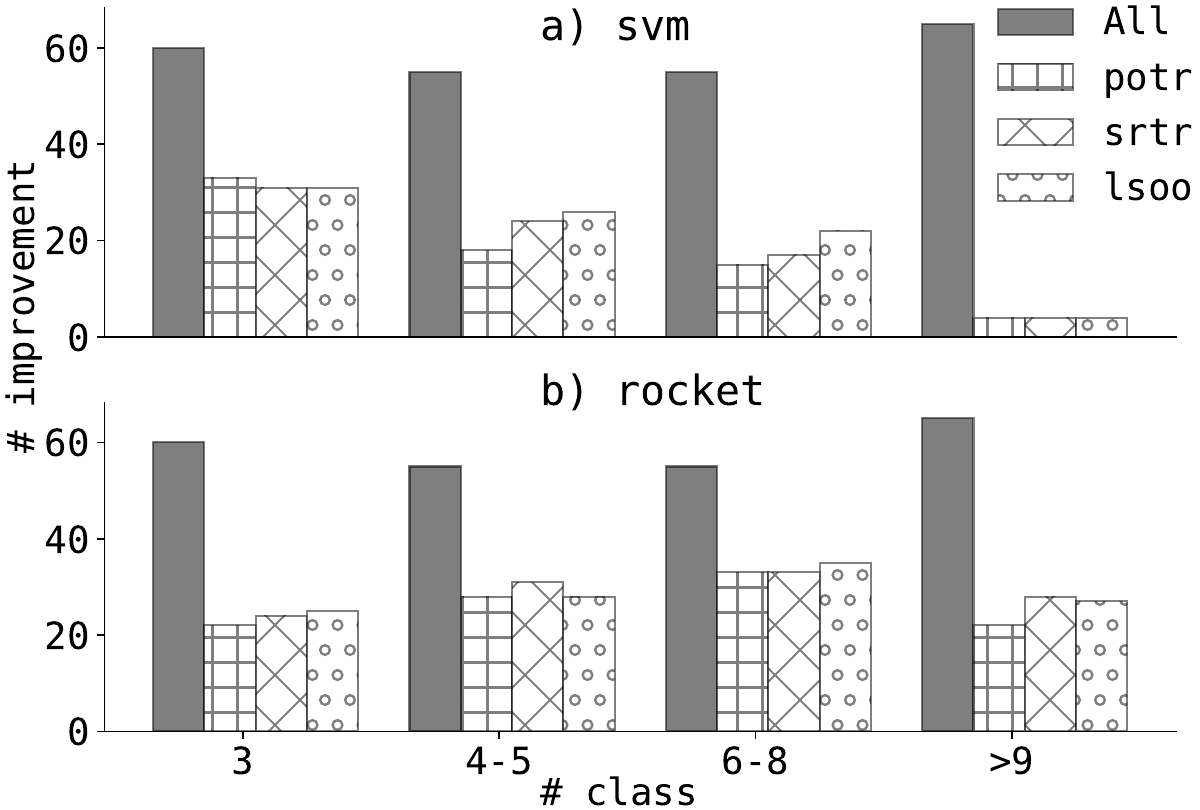}
  \caption{The figure illustrates a comparison of the number of classes versus the number of improved results for \texttt{svm} (a) and \texttt{rocket} (b) when utilizing different splitting functions, namely \texttt{potr}, \texttt{srtr}, and \texttt{lsoo}. The tests were performed with 10 iterations, considering the individual testing scores of the nested CV scheme. To ensure a balanced representation of the total observations in each bin of the grouped bar chart, some classes were grouped together.}
  \hypertarget{fig:nclass}{}
  \label{fig:nclass}
\end{figure}

In order to have a comprehensive visual overview of the data distribution and relationships between the features and the outcome, Figure \hyperlink{fig:pcloud}{4} presents a visual comparison between the remaining dataset features and the improvement obtained using HC over FC, represented as a point cloud. In the case of \texttt{rocket}, the dark-colored markers are concentrated more towards the bottom of the point cloud space, which aligns with the negative correlation coefficient identified through the Pearson correlation test. However, for the other features, no significant concentration or patterns are observed in the point cloud sets. This indicates that there may not be strong linear relationships between these features and the improvement obtained using HC over FC.
\begin{figure*}[ht]
  \centering
  \includegraphics[width=\textwidth]{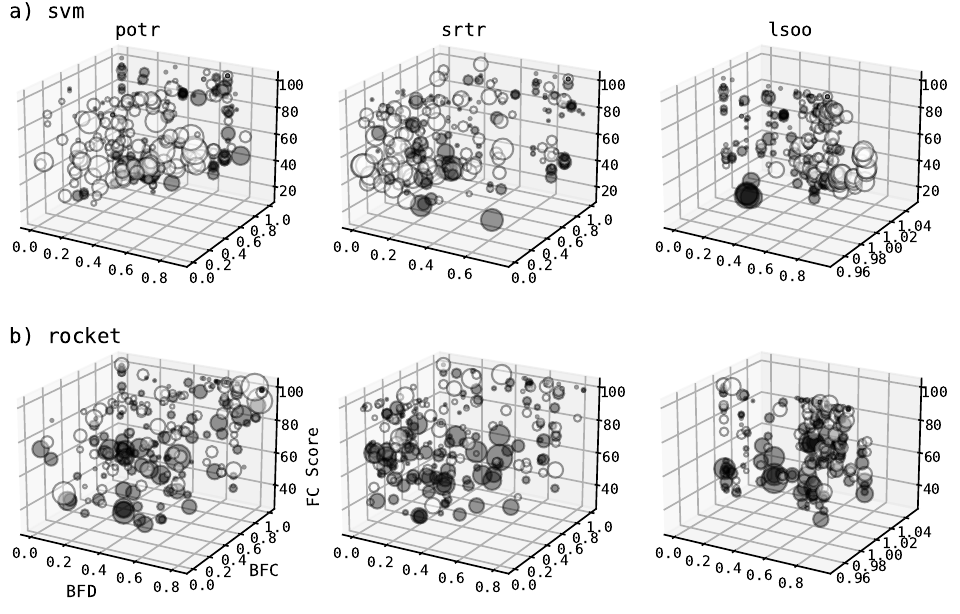}
  \caption{Point cloud visualization of dataset features extracted using the classifiers \texttt{svm} (a) and \texttt{rocket} (b) and SSFs \texttt{potr}, \texttt{srtr}, and \texttt{lsoo} as indicated (on the left, center, and right, respectively). The tests were conducted with 10 iterations, taking into account the individual testing scores of the nested CV scheme. The x-axis and y-axis of the point cloud represent BFD and BFC, respectively. The z-axis shows FC score. The marker color indicates the improvement state; dark color shows there is an improvement, and light color shows there is no improvement. The marker size represents the difference of performance evaluation between HC and FC, denoted as $\Delta g$; a larger marker size indicates a larger $\Delta g$.}
  \hypertarget{fig:pcloud}{}
  \label{fig:pcloud}
\end{figure*}

Figures \hyperlink{fig:bfc_vs_delta}{5}, \hyperlink{fig:bfd_vs_delta}{6}, and \hyperlink{fig:fc_vs_delta}{7} in the Appendix present graphical representations of the comparison between each individual feature (i.e., BFC, BFD, and FC score) and $\Delta g$, which represents the difference between HC and FC scores, using 2D point clouds with fitted linear regression lines.

In Figure \hyperlink{fig:bfc_vs_delta}{5}, showing the relationship between BFC and $\Delta g$, a positive-sloped line is observed for \texttt{svm} in both \texttt{potr} and \texttt{srtr} cases, while a flattish line is observed for all cases of \texttt{rocket}.

In Figure \hyperlink{fig:bfd_vs_delta}{6}, illustrating the relationship between BFD and $\Delta g$, all cases exhibit flattish lines except for \texttt{svm}, where a positive-sloped line is observed for \texttt{srtr} and a negative-sloped line is observed for \texttt{lsoo}.

Lastly, in Figure \hyperlink{fig:fc_vs_delta}{7}, displaying the relationship between FC score and $\Delta g$, all lines are found to be flattish for \texttt{svm}, while they are negatively sloped for \texttt{rocket}.

These graphical representations provide visual confirmation of the findings obtained from the Pearson correlation tests, supporting the presence of certain relationships between the features and the improvement achieved using HC over FC.

\section{Discussions}
\subsection{Discussions of the Results}
The results obtained from this study offer valuable insights into the effectiveness and behavior of the proposed hierarchical divisive clustering approach with SSFs for enhancing classification performance in multi-class datasets. The comparison between the classifiers \texttt{svm} and \texttt{rocket} indicates that \texttt{rocket} consistently outperforms \texttt{svm} in terms of the number of improved datasets. This observation suggests that \texttt{rocket} is better equipped to handle the hierarchical data structure of time series data. The advantage of \texttt{rocket} can be attributed to its specialized design tailored for time series data, enabling it to capture temporal dependencies and patterns effectively. As a result, it achieves enhanced classification performance in the hierarchical divisive clustering, which strongly relies on the performance of the underlying classifier.

Conversely, \texttt{svm}, being a more generic classifier, may not fully capitalize on the time series-specific information, leading to comparatively fewer improvements in this hierarchical context. The observed trend further highlights the importance of using domain-specific classifiers for time series datasets, where the inherent temporal nature of the data can significantly impact the classification results. The findings underscore the benefits of employing specialized classifiers like \texttt{rocket} when dealing with time series data in hierarchical clustering scenarios.

Furthermore, the flat CV scheme demonstrated a higher number of improvements compared to the nested CV scheme, as anticipated due to the increased optimization opportunities provided by the former. Specifically, in the flat CV scheme, an inherent bias was introduced during the evaluation phase when searching for the optimal tree. This observation provides valuable insight, indicating that there is potential to enhance the generalization performance of the proposed approach. Although some trees were generated during the nested CV scheme, they escaped notice during the evaluation phase. This realization serves as motivation to further refine and improve the technique for finding the optimal tree in the hierarchical divisive clustering process.

Regarding the number of iterations in the hierarchical divisive clustering process, it was found that using 3, 5, or 10 iterations is sufficient to achieve meaningful improvements for both \texttt{rocket} and \texttt{svm}. The number of improvements declined when the number of iterations exceeded 20. Therefore, the recommendation is to use 10 iterations for \texttt{svm} to strike a balance between improvement and computational cost.

The correlation and regression analyses provide additional insights into the relationship between dataset features and the improvements achieved through hierarchical divisive clustering. The results consistently demonstrate that the number of classes and FC score are significant features for \texttt{svm} and \texttt{rocket}, respectively. However, when it comes to BFC and BFD metrics, there were discrepancies in their significance depending on the classifier and SSF used, as well as the specific statistical tests applied. These variations may be attributed to the intricate relationships between the dataset features and the hierarchical structure, as well as the specific characteristics of each classifier and splitting function. 

The findings from Tables \hyperlink{tab:svm}{2} and \hyperlink{tab:rocket}{3} highlight an interesting pattern: for datasets with a smaller number of classes, consistently better results are observed when imposing a hierarchy over FC, particularly in cases involving at least one of the SSFs. This observation suggests that for datasets with fewer classes, there may exist at least one hierarchical structure that outperforms the FC approach.

One possible explanation for this pattern is the reduced search space for trees when dealing with datasets with a smaller number of classes. With fewer classes, the number of possible hierarchical structures to consider is smaller, making it more feasible to explore the space of potential hierarchies and identify ones that lead to improved performance.

As a result, the focus of future works should be on improving the efficiency and efficacy of the proposed approach in finding these hierarchical structures, even for datasets with a larger number of classes. Developing more efficient algorithms or optimization techniques that can effectively search through the larger search space of trees could lead to the discovery of valuable hierarchical structures that enhance the classification performance.

\subsection{Limitations of the Approach}
Indeed, the stochastic nature of the approach, while beneficial in enhancing the exploration aspect, also introduces certain limitations. These limitations can be summarized as follows: i) Efficiency: The proposed approach requires multiple iterations, which can decrease efficiency compared to deterministic methods that converge in fewer iterations. ii) Convergence: Convergence of the proposed approach is not guaranteed, and the convergence behavior can be erratic. It may take a varying number of iterations for the algorithm to converge, making it challenging to determine when the optimal solution is reached. iii) Solution Quality: The proposed approach does not guarantee finding the true solution. The obtained solution may only be a local optimum or a suboptimal solution, depending on the specific problem and the randomness introduced during the iterations. iv) Sampling Bias: The stochastic nature of the approach introduces sampling bias, which can affect the generalization performance of the algorithm. The sampled subset may not be representative of the entire dataset, leading to suboptimal performance on unseen data.

To address this limitation, future research can focus on implementing advanced optimization techniques that strike a balance between exploration and exploitation. Techniques such as adaptive learning rates, dynamic exploration probabilities, or guided search strategies can be explored to optimize the trade-off between stochasticity and convergence.

An additional noteworthy limitation of the approach is its computational cost, which can be up to approximately $N_{iter}|C|^2$ times more expensive than FC (as discussed in Section \hyperlink{sec:cost}{3.3}). Nonetheless, the impact of this cost on the overall efficiency can be moderated by utilizing multiprocessing techniques. Moreover, the implementation of the proposed approach is not impractical, as it is relatively easy to implement. The algorithms and techniques used in the approach, such as the SSFs and hierarchical divisive clustering, can be implemented using standard programming libraries and tools. The source code\footnotemark[\value{footnote}] and detailed explanations provided in the study's materials can further aid in the implementation process. This simplicity in implementation makes the approach accessible and user-friendly for researchers and practitioners interested in exploring HC methods.

\subsection{Future Works}
The proposed hierarchical divisive clustering approach with SSFs offers a promising framework for improving classification performance in multi-class time series datasets. However, there are several avenues for future exploration and improvement:

\paragraph{Improving Efficacy, Generalizability, and Efficiency:} Exploring additional dataset features and metrics, as well as novel splitting functions, could provide deeper insights into their impact on clustering performance. Incorporating a wider range of classifiers specifically tailored for time series data and exploring ensemble techniques could enhance the generalizability of the approach. Efforts to reduce computational cost through more efficient algorithms and parallelization techniques will be essential to scale the approach to larger datasets and make it practically applicable in various domains.

\paragraph{Exploring Different Hierarchy Structures:} Currently, the approach follows a Local Classifier per Node (LCPN) configuration, but it can be adapted to different hierarchy setups, such as a global approach. However, incorporating a global approach for analyzing time series data comes with challenges, especially when adapting traditional algorithms not explicitly designed for hierarchical structures. Exploring decision trees or uncertainty forests as base classifiers may be potential solutions to effectively handle class hierarchies and temporal dependencies in time series data.

\paragraph{Applying the Approach to Different Domains:} The proposed approach can be extended to various domains and applications beyond time series data. Comparing its performance with state-of-the-art hierarchical clustering techniques in different domains will provide valuable insights into its effectiveness and competitiveness.

In summary, future research should focus on refining the efficacy, generalizability, and efficiency of the proposed approach by exploring new features, classifiers, and optimization techniques. Additionally, investigating different hierarchy structures and adapting the approach to various domains will further expand its applicability and potential impact in real-world scenarios. With continuous efforts and advancements, the proposed approach can become a valuable tool for hierarchical clustering and classification tasks in diverse fields.

\section{Conclusions}
In this study, a novel approach was presented for enhancing HC by generating the hierarchy of classes from a given set of flat classes, without requiring explicit hierarchical information. The proposed method employed a hierarchical divisive clustering approach with SSFs, which systematically divided the set of classes into two subsets based on their similarity in terms of discriminability by the classifier. This allowed for the construction of a binary tree representation of hierarchical classes that maximized the classification score between the formed groups.

The efficacy of the approach was evaluated on a diverse collection of 46 multi-class datasets from the UCR archive. Two popular classifiers, \texttt{svm} and \texttt{rocket}, and three SSFs, \texttt{potr}, \texttt{srtr}, and \texttt{lsoo}, were utilized to assess the impact on HC performance.

The results demonstrated that the proposed method significantly improved HC performance, particularly when \texttt{rocket} was used as the classifier. This finding underscored the importance of effectively leveraging the hierarchical structure when dealing with complex and diverse datasets. Furthermore, the use of the flat CV scheme led to more improvements compared to the nested CV scheme, indicating the potential of the approach in optimizing hierarchical classifiers.

A key advantage of the approach was its ability to generate the hierarchy of classes when explicit hierarchical information was not available. By automatically partitioning the classes into two subsets based on their similarity, the approach offered a systematic and efficient solution for constructing the hierarchical tree. Additionally, the concepts of BFC and BFD were introduced to characterize the balance within the hierarchical tree structure, providing valuable insights for further analysis.

The study also investigated the relationship between dataset features and the improvement in HC. While the number of classes and FC score were found to be significant factors for \texttt{svm} and \texttt{rocket}, respectively, some variations in the results were observed depending on the choice of splitting function.

In conclusion, the proposed hierarchical divisive clustering approach with SSFs offers an effective strategy for enhancing HC, particularly in cases where explicit hierarchy information is not provided. By optimizing the hierarchical tree structure and effectively utilizing the advantages of the hierarchical organization, the approach shows promising potential in improving classification performance. Future research can further explore different splitting functions and classifiers to achieve even better performance in HC tasks.



\normalsize
\bibliography{hc4tsc_hdc_ssf}

\onecolumn
\section{Appendix}
\subsection{Deriving Number of Datapoints to be Processed}
\paragraph{During preprocessing and training phase:} Let $O_{X}$ denote the number of datapoints to be processed. 

Assuming each parent node splits into two subsets both with equal number of datapoints and equal number of classes, namely $BFC(P)=BFD(P,X)=0$:
\begin{align}
\begin{split}
O_{X}& = |X_{tr}||C|+2\frac{|X_{tr}|}{2}\frac{|C|}{2}+2^2\frac{|X_{tr}|}{2^2}\frac{|C|}{2^2}+...+2^{log_2|C|\text{-}1}\frac{|X_{tr}|}{2^{log_2|C|\text{-}1}}\frac{|C|}{2^{log_2|C|\text{-}1}} \\
& = |X_{tr}||C|(2^0+2^{\text{-}1}+...+2^{\text{-}log_2|C|+1}) \\
& = |X_{tr}||C| \sum_{k=0}^{log_2|C|\text{-}1}2^{\text{-}k} \\
& \approx 2|X_{tr}||C| \\
\end{split}
\end{align}
On the contrary, assuming each parent node splits into two subsets with maximum imbalance where the difference between the subsets in both the number of datapoints and the number of classes is the greatest, namely $BFC(P)=BFD(P,X)=\{\text{-}1,1\}$.:
\begin{align}
\begin{split}
O_{X} & = |X_{tr}||C|+(|X_{tr}|\text{-}1)(|C|\text{-}1)+...+(|X_{tr}|\text{-}|C|+1)(|C|\text{-}|C|+1) \\
& = |X_{tr}||C| + |X_{tr}||C| \text{-} (|X_{tr}|+|C|)+1+...+|X_{tr}||C| \text{-} (|C|\text{-}1)(|X_{tr}|+|C|)+(|C|\text{-}1)^2 \\
& = |X_{tr}||C|(|C|\text{-}1) \text{-} (|X_{tr}|+|C|)\sum_{k=1}^{|C|\text{-}1}k + \sum_{k=1}^{|C|\text{-}1}k^2 \\
& = |X_{tr}||C|(|C|\text{-}1) \text{-} \frac{1}{2}(|X_{tr}|+|C|)|C|(|C|\text{-}1) + \frac{1}{6}(2|C|\text{-}1)|C|(|C|\text{-}1) \\
& = \frac{1}{6}(3|X_{tr}||C|^2 \text{-} 3|X_{tr}||C| \text{-} |C|^3 + |C|) \\
& \approx \frac{1}{2}|X_{tr}||C|^2 \\
\end{split}
\end{align}
where the analysis is simplified ignoring the lower order terms. Since number of datapoints is typically much larger than the number of classes, $|X_{tr}|>>|C|$, then the term $|C|^3$ will have negligible effect on the outcome and it is ignored too.




\paragraph{During testing phase:} Assuming maximum imbalance of class labels at parent nodes while maintaining an even distribution of data instances among the different classes, the average tree depth can be approximated as:
\begin{align}
\begin{split}
        & \frac{1}{|C|} (1 + 2 +...+ |C| \text{-}1) + |C| \text{-}1)\\ 
=       & \frac{1}{|C|} (\sum_{i=1}^{|C|\text{-}1} i + |C| \text{-}1)\\ 
=       & \frac{|C|}{2} \text{-} \frac{1}{|C|} \text{-} \frac{5}{2}\\
\approx & |C|/2\\
\end{split}
\end{align}

Please note that when considering the balance in the data distribution of different classes, the average tree depth value can be approximated between 1 and the number of classes, denoted as $|C|$. Specifically, if the classes that are separated around the top of the tree have more weight, the average tree depth value will tend to be closer to 1. Conversely, if the classes that are separated around the bottom of the tree have more weight, the average tree depth value will tend to be closer to $|C|$. 

As other extreme for the average tree depth where maximum balance of tree is assumed, following approximation entails:
\begin{equation}
log_{2}|C|
\end{equation}

\subsection{Supplementary Figures, Tables}
\begin{table*}[t!]
 \caption{\texttt{svm} results with $N_{iter}=10$ iterations. nCV and fCV stand for nested and flat CV scores, respectively.}\label{tab:svm}
 \hypertarget{tab:svm}{}
  \centering
  \begin{tabular}{lrccccccc}
    \toprule
    \multicolumn{3}{c}{Datasets} & \multicolumn{2}{c}{\texttt{potr}} & \multicolumn{2}{c}{\texttt{srtr}} & \multicolumn{2}{c}{\texttt{lsoo}} \\
    \cmidrule(r){1-3} \cmidrule(r){4-5} \cmidrule(r){6-7} \cmidrule(r){8-9}
    Name                       & $|C|$ & FC  & nCV & fCV & nCV & fCV & nCV & fCV\\
    \midrule
    Adiac&37&62.42&55.95&58.44&59.43&61.88&48.19&51.11\\
    ArrowHead&3&86.89&85.19&\textbf{87.60}&85.19&86.71&85.19&\textbf{87.60}\\
    Beef&5&84.06&81.92&\textbf{93.33}&81.70&\textbf{88.67}&83.07&\textbf{91.73}\\
    Car&4&83.45&75.63&\textbf{86.68}&80.91&\textbf{84.42}&81.69&\textbf{84.17}\\
    ChlorineConcentration&3&80.91&\textbf{84.06}&\textbf{84.06}&\textbf{84.06}&\textbf{84.06}&\textbf{84.06}&\textbf{84.06}\\
    CricketX&12&42.16&29.11&33.23&29.00&32.73&29.55&32.43\\
    CricketY&12&43.41&31.27&35.46&31.49&34.56&33.62&37.55\\
    CricketZ&12&38.64&28.75&30.44&26.64&31.82&27.51&31.88\\    DiatomSizeReduction&4&98.97&\textbf{99.22}&\textbf{99.49}&\textbf{99.49}&\textbf{99.49}&\textbf{99.22}&\textbf{100.00}\\    DistalPhalanxOutlineAgeGroup&3&68.58&\textbf{73.27}&\textbf{73.27}&\textbf{70.08}&\textbf{70.08}&\textbf{70.08}&\textbf{70.08}\\    DistalPhalanxTW&6&51.73&50.34&\textbf{53.87}&\textbf{54.15}&\textbf{56.56}&\textbf{53.15}&\textbf{54.00}\\
    ECG5000&5&60.63&58.62&\textbf{63.72}&60.19&\textbf{63.85}&60.08&\textbf{61.18}\\
    FaceAll&14&89.08&81.71&82.74&81.90&83.68&82.82&84.88\\    FaceFour&4&93.54&\textbf{95.16}&\textbf{96.17}&\textbf{96.00}&\textbf{97.05}&\textbf{96.00}&\textbf{97.05}\\
    FacesUCR&14&90.71&83.95&86.95&84.22&85.82&83.71&85.88\\
    FiftyWords&50&52.71&26.13&33.20&28.37&32.72&35.85&39.38\\
    Fish&7&86.78&81.63&\textbf{88.94}&81.83&85.95&85.29&\textbf{88.44}\\
    Haptics&5&44.99&40.74&\textbf{47.47}&\textbf{45.19}&\textbf{46.71}&44.10&\textbf{49.01}\\
    InlineSkate&7&37.89&\textbf{38.16}&\textbf{42.89}&\textbf{39.90}&\textbf{41.34}&33.81&35.87\\
    InsectWingbeatSound&11&65.11&64.55&\textbf{65.51}&63.99&\textbf{65.56}&63.51&\textbf{66.03}\\    LargeKitchenAppliances&3&37.27&\textbf{39.12}&\textbf{39.84}&\textbf{39.12}&\textbf{39.84}&\textbf{39.12}&\textbf{39.84}\\
    Lightning7&7&56.70&52.68&\textbf{61.22}&\textbf{58.50}&\textbf{64.59}&55.35&\textbf{61.35}\\
    Mallat&8&98.41&98.37&98.41&\textbf{98.42}&\textbf{98.42}&98.41&98.41\\
    Meat&3&97.47&97.47&\textbf{100.00}&95.78&96.61&97.47&\textbf{98.31}\\
    MedicalImages&10&61.87&58.46&60.60&60.14&\textbf{62.18}&60.45&\textbf{62.13}\\    MiddlePhalanxOutlineAgeGroup&3&64.11&\textbf{66.81}&\textbf{67.34}&\textbf{64.26}&\textbf{64.26}&\textbf{64.26}&\textbf{64.26}\\
    MiddlePhalanxTW&6&38.98&38.46&38.97&37.96&\textbf{39.38}&37.66&38.34\\
    OliveOil&4&17.59&\textbf{39.34}&\textbf{39.46}&\textbf{38.64}&\textbf{43.02}&\textbf{38.55}&\textbf{43.48}\\
    OSULeaf&6&46.47&42.10&\textbf{47.05}&44.07&\textbf{48.05}&42.08&\textbf{47.88}\\
    ProximalPhalanxOutlineAgeGroup&3&80.09&79.09&\textbf{81.06}&80.09&\textbf{80.30}&80.09&80.09\\
    ProximalPhalanxTW&6&44.65&43.41&\textbf{45.83}&44.02&\textbf{45.62}&\textbf{44.89}&\textbf{45.72}\\
    RefrigerationDevices&3&37.72&35.45&\textbf{37.86}&35.45&\textbf{37.86}&35.45&\textbf{37.86}\\    ScreenType&3&32.88&\textbf{33.76}&\textbf{36.11}&\textbf{33.76}&\textbf{36.11}&\textbf{33.76}&\textbf{36.11}\\    SmallKitchenAppliances&3&42.77&\textbf{44.85}&\textbf{45.45}&\textbf{43.50}&\textbf{45.45}&\textbf{43.50}&\textbf{45.45}\\
    SwedishLeaf&15&86.98&81.15&84.58&83.90&85.20&82.21&83.81\\
    Symbols&6&95.71&95.31&\textbf{96.20}&95.50&95.71&\textbf{95.80}&\textbf{95.80}\\
    SyntheticControl&6&94.66&94.64&\textbf{95.30}&93.81&\textbf{94.83}&93.64&\textbf{94.82}\\
    WordSynonyms&25&42.82&26.48&31.77&26.41&31.50&32.35&35.37\\
    Worms&5&44.18&40.04&\textbf{46.37}&43.71&\textbf{47.51}&42.95&\textbf{48.12}\\
    ACSF1&10&58.44&50.31&57.00&48.51&\textbf{60.10}&54.62&\textbf{58.99}\\    EthanolLevel&4&73.71&\textbf{74.27}&\textbf{75.46}&\textbf{74.13}&\textbf{74.52}&\textbf{74.85}&\textbf{75.15}\\
    Rock&4&87.13&\textbf{87.18}&\textbf{93.27}&\textbf{87.18}&\textbf{92.18}&87.13&\textbf{92.18}\\
    SemgHandMovementCh2&6&62.01&58.76&61.17&58.18&60.12&60.13&60.13\\
    SemgHandSubjectCh2&5&88.19&85.02&86.52&85.38&87.49&86.60&87.12\\    SmoothSubspace&3&90.16&\textbf{90.47}&\textbf{90.91}&\textbf{90.47}&\textbf{90.47}&\textbf{90.47}&\textbf{90.47}\\
    UMD&3&98.36&98.36&\textbf{98.88}&98.36&\textbf{98.88}&98.36&\textbf{98.88}\\
    \bottomrule
  \end{tabular}
  \label{tab:table}
\end{table*}

\begin{table*}[t!]
 \caption{\texttt{rocket} results with $N_{iter}=10$ iterations. nCV and fCV stand for nested and flat CV scores, respectively.}\label{tab:rocket}
 \hypertarget{tab:rocket}{}
  \centering
  \begin{tabular}{lrccccccc}
    \toprule
    \multicolumn{3}{c}{Datasets} & \multicolumn{2}{c}{\texttt{potr}} & \multicolumn{2}{c}{\texttt{srtr}} & \multicolumn{2}{c}{\texttt{lsoo}} \\
    \cmidrule(r){1-3} \cmidrule(r){4-5} \cmidrule(r){6-7} \cmidrule(r){8-9}
    Name                       & $|C|$ & FC  & nCV & fCV & nCV & fCV & nCV & fCV\\
    \midrule
Adiac&37&75.93&\textbf{77.01}&\textbf{79.11}&74.36&\textbf{79.21}&\textbf{77.05}&\textbf{79.16}\\
ArrowHead&3&93.89&92.42&\textbf{93.90}&92.42&\textbf{93.90}&92.42&\textbf{93.90}\\
Beef&5&70.97&61.32&\textbf{80.71}&\textbf{71.87}&\textbf{78.71}&69.70&\textbf{83.82}\\
Car&4&87.29&\textbf{91.55}&\textbf{94.94}&\textbf{93.23}&\textbf{94.94}&\textbf{93.23}&\textbf{94.94}\\
ChlorineConcentration&3&98.06&97.75&\textbf{98.83}&97.75&\textbf{98.83}&97.75&\textbf{98.83}\\
CricketX&12&76.77&\textbf{78.55}&\textbf{80.77}&\textbf{78.72}&\textbf{80.35}&\textbf{77.49}&\textbf{79.54}\\
CricketY&12&74.48&73.48&\textbf{76.37}&73.47&\textbf{77.42}&71.35&74.10\\
CricketZ&12&74.36&\textbf{75.58}&\textbf{77.64}&\textbf{75.90}&\textbf{77.55}&\textbf{74.43}&\textbf{76.74}\\
DiatomSizeReduction&4&100.00&98.91&100.00&99.48&100.00&100.00&100.00\\
DistalPhalanxOutlineAgeGroup&3&77.98&\textbf{78.99}&\textbf{79.56}&\textbf{78.99}&\textbf{79.48}&\textbf{78.99}&\textbf{79.48}\\
DistalPhalanxTW&6&49.77&\textbf{52.24}&\textbf{56.49}&\textbf{53.36}&\textbf{56.71}&\textbf{54.47}&\textbf{55.40}\\
ECG5000&5&61.86&\textbf{62.09}&\textbf{69.44}&\textbf{62.65}&\textbf{65.18}&\textbf{64.51}&\textbf{66.22}\\
FaceAll&14&97.48&95.14&96.88&95.45&96.77&96.10&96.94\\
FaceFour&4&96.68&92.35&\textbf{97.48}&92.20&\textbf{97.48}&94.85&\textbf{97.48}\\
FacesUCR&14&98.54&97.73&98.40&97.26&98.20&97.09&98.33\\
FiftyWords&50&56.24&\textbf{56.74}&\textbf{61.27}&54.95&\textbf{62.54}&\textbf{61.07}&\textbf{64.86}\\
Fish&7&79.06&\textbf{92.01}&\textbf{95.74}&\textbf{90.86}&\textbf{94.89}&\textbf{91.36}&\textbf{93.36}\\
Haptics&5&52.86&50.77&\textbf{56.67}&48.67&\textbf{56.10}&49.49&\textbf{53.80}\\
InlineSkate&7&69.61&67.67&\textbf{70.28}&67.39&\textbf{71.03}&68.54&\textbf{70.70}\\
InsectWingbeatSound&11&68.61&68.41&\textbf{70.36}&\textbf{69.04}&\textbf{70.22}&68.21&\textbf{69.62}\\
LargeKitchenAppliances&3&83.59&81.37&\textbf{84.30}&81.37&\textbf{83.98}&81.64&\textbf{83.90}\\
Lightning7&7&75.53&\textbf{79.82}&\textbf{86.02}&\textbf{77.09}&\textbf{85.03}&\textbf{78.14}&\textbf{83.41}\\
Mallat&8&99.50&99.37&\textbf{99.75}&99.29&\textbf{99.62}&99.33&99.46\\
Meat&3&99.16&95.80&99.16&95.80&99.16&95.80&99.16\\
MedicalImages&10&77.80&75.98&\textbf{79.91}&76.13&\textbf{79.99}&\textbf{78.46}&\textbf{80.25}\\
MiddlePhalanxOutlineAgeGroup&3&64.57&\textbf{65.44}&\textbf{66.68}&64.46&64.46&64.46&64.46\\
MiddlePhalanxTW&6&37.75&\textbf{39.01}&\textbf{40.07}&\textbf{37.90}&\textbf{40.81}&37.32&\textbf{38.47}\\
OliveOil&4&93.94&86.91&\textbf{98.73}&91.53&\textbf{98.73}&90.40&\textbf{97.46}\\
OSULeaf&6&87.21&\textbf{88.16}&\textbf{89.92}&\textbf{87.55}&\textbf{89.37}&\textbf{87.25}&\textbf{90.88}\\
ProximalPhalanxOutlineAgeGroup&3&79.99&77.67&\textbf{81.73}&79.83&79.83&79.83&79.83\\
ProximalPhalanxTW&6&45.80&\textbf{50.93}&\textbf{56.10}&\textbf{49.23}&\textbf{55.91}&\textbf{50.07}&\textbf{53.20}\\
RefrigerationDevices&3&47.35&46.37&\textbf{53.93}&46.37&\textbf{53.93}&46.37&\textbf{52.92}\\
ScreenType&3&41.46&39.48&\textbf{43.51}&\textbf{41.64}&\textbf{43.51}&\textbf{41.57}&\textbf{43.51}\\
SmallKitchenAppliances&3&69.24&68.48&\textbf{69.98}&67.48&67.96&67.89&67.89\\
SwedishLeaf&15&94.54&\textbf{94.63}&\textbf{96.09}&\textbf{95.39}&\textbf{96.00}&\textbf{94.84}&\textbf{96.37}\\
Symbols&6&98.43&\textbf{98.73}&\textbf{99.22}&\textbf{98.73}&\textbf{98.93}&\textbf{98.53}&\textbf{98.83}\\
SyntheticControl&6&99.67&99.50&\textbf{100.00}&99.67&\textbf{100.00}&99.33&\textbf{100.00}\\
WordSynonyms&25&61.10&56.30&\textbf{65.57}&58.21&\textbf{63.89}&60.60&\textbf{65.31}\\
Worms&5&50.53&\textbf{59.57}&\textbf{62.29}&\textbf{58.72}&\textbf{63.63}&\textbf{57.24}&\textbf{63.08}\\
ACSF1&10&65.50&\textbf{69.86}&\textbf{76.72}&\textbf{73.74}&\textbf{76.50}&\textbf{72.34}&\textbf{78.22}\\
EthanolLevel&4&52.27&\textbf{71.19}&\textbf{73.13}&\textbf{72.44}&\textbf{73.56}&\textbf{72.44}&\textbf{72.55}\\
Rock&4&88.94&\textbf{91.76}&\textbf{91.76}&88.84&\textbf{91.76}&88.84&\textbf{91.34}\\
SemgHandMovementCh2&6&35.66&\textbf{40.24}&\textbf{45.01}&\textbf{40.57}&\textbf{45.23}&\textbf{39.29}&\textbf{42.43}\\
SemgHandSubjectCh2&5&48.75&\textbf{55.95}&\textbf{58.33}&\textbf{56.91}&\textbf{59.07}&\textbf{58.13}&\textbf{59.50}\\
SmoothSubspace&3&97.00&96.97&\textbf{98.99}&96.97&\textbf{98.99}&96.97&\textbf{98.32}\\
UMD&3&98.30&\textbf{98.88}&\textbf{99.44}&\textbf{98.88}&\textbf{99.44}&\textbf{98.88}&\textbf{99.44}\\   
    \bottomrule
  \end{tabular}
  \label{tab:table}
\end{table*}

\begin{figure*}[ht]
  \centering
  \includegraphics[width=0.9\textwidth]{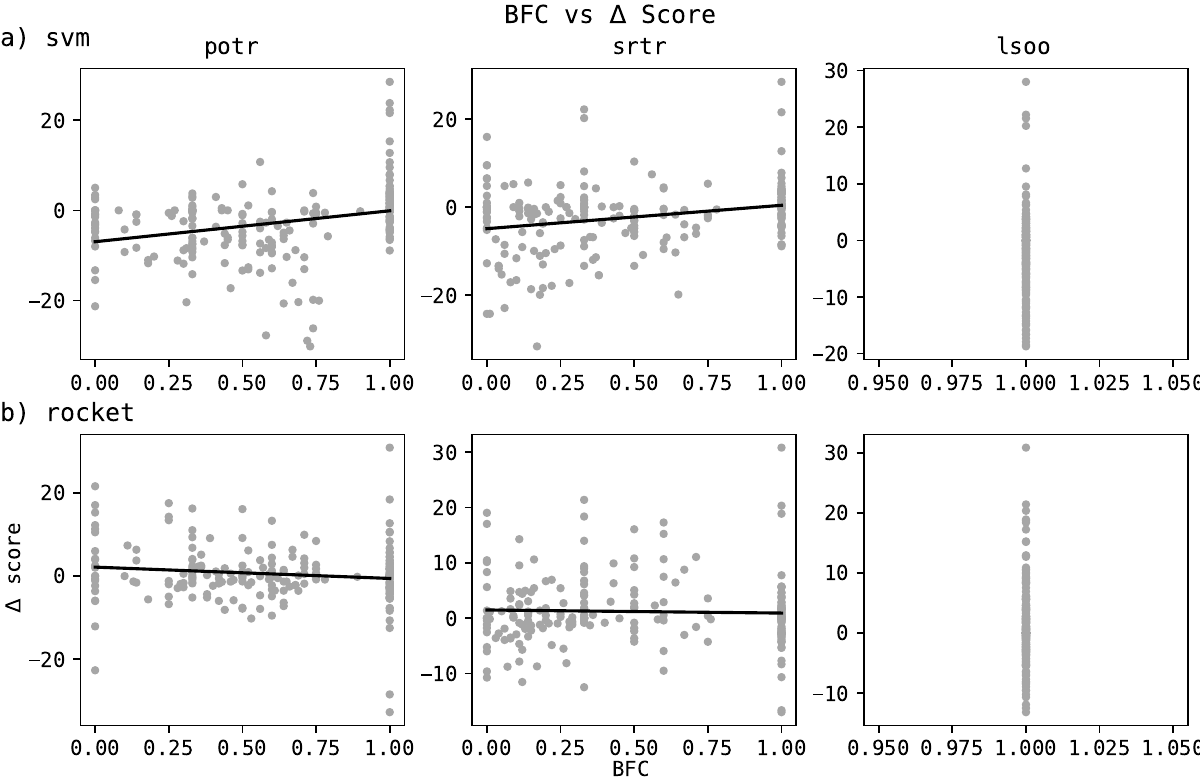}
  \caption{Point cloud visualization of the relationship between BFC and $\Delta g$ (difference between HC and FC score) in case of using \texttt{svm} (a) and \texttt{rocket} (b) as the classifier and using \texttt{potr} (on the left), \texttt{srtr} (on the center), \texttt{lsoo} (on the right) as the SSF. Each point in the point cloud represents an observation from the testing phase of the nested cross-validation scheme for the respective case, with five observations per dataset. The x-axis of the point cloud shows the BFC values, indicating the balance or imbalance between class labels in the hierarchical tree structure. The y-axis represents $\Delta g$, revealing the difference in classification performance between HC and FC. Indeed, because of how \texttt{lsoo} works, BFC in that case is strictly 1. For each combination of classifier and splitting function, a regression line is fitted to the point cloud via linear regression. This gives insight into the overall trend or relationship between BFC and $\Delta g$ to be assessed. The tests were conducted with $N_{iter}=10$ iterations.}
  \hypertarget{fig:bfc_vs_delta}{}
  \label{fig:bfc_vs_delta}
\end{figure*}

\begin{figure*}[ht]
  \centering
  \includegraphics[width=0.9\textwidth]{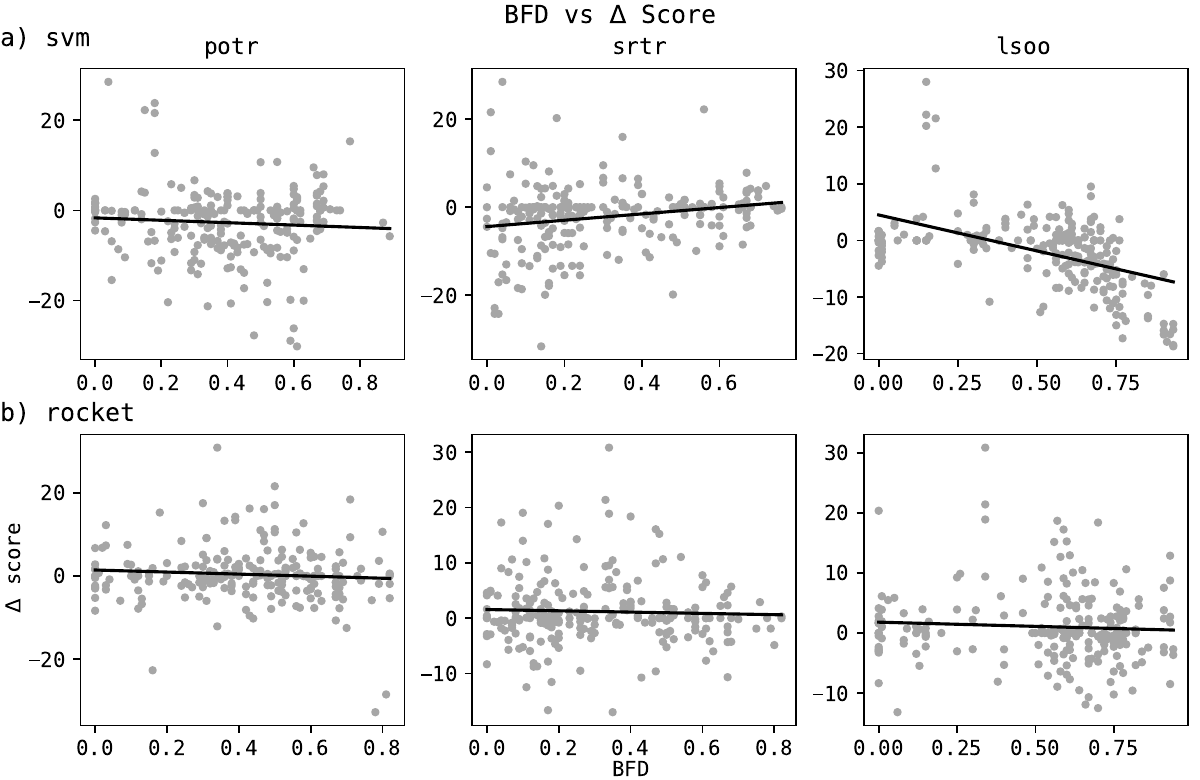}
  \caption{The point cloud visualization illustrates the correlation between BFD and $\Delta g$ (the difference between HC and FC scores) in two cases: when using \texttt{svm} as the classifier (a) and when using \texttt{rocket} as the classifier (b). The SSFs \texttt{potr}, \texttt{srtr}, and \texttt{lsoo} are utilized as indicated (on the left, center, and right, respectively). Each point in the point cloud corresponds to an observation obtained during the testing phase of the nested CV scheme for each specific scenario, with five observations per dataset. The x-axis of the point cloud represents the BFD values, which quantify the balance or imbalance between class labels in the hierarchical tree structure. The y-axis reflects $\Delta g$, revealing the difference in classification performance between HC and FC. To assess the overall trend or relationship between BFD and $\Delta g$, a regression line is fitted to the point cloud via linear regression for each combination of classifier and splitting function. This analysis provides insights into the association between BFD and the performance difference between hierarchical and FC schemes. The tests were conducted with $N_{iter}=10$ iterations.}
  \hypertarget{fig:bfd_vs_delta}{}
  \label{fig:bfd_vs_delta}
\end{figure*}

\begin{figure*}[ht]
  \centering
  \includegraphics[width=0.9\textwidth]{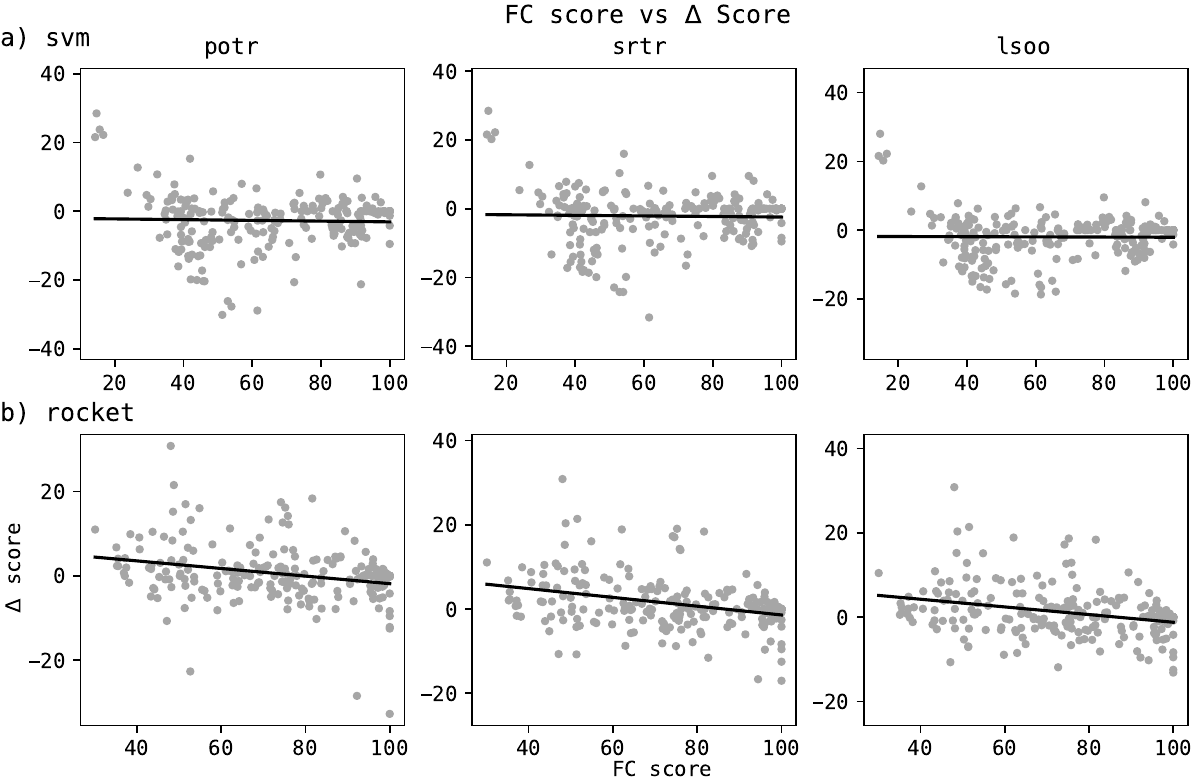}
  \caption{The point cloud visualization illustrates the correlation between FC score and the difference between HC and FC scores, denoted as $\Delta g$, in two cases: when using \texttt{svm} as the classifier (a) and when using \texttt{rocket} as the classifier (b). The SSFs \texttt{potr}, \texttt{srtr}, and \texttt{lsoo} are utilized as indicated (on the left, center, and right, respectively). Each point in the point cloud corresponds to an observation obtained during the testing phase of the nested CV scheme for each specific scenario, with five observations per dataset. The x-axis of the point cloud represents the FC scores. The y-axis reflects $\Delta g$, revealing the difference in classification performance between HC and FC. To assess the overall trend or relationship between FC score and $\Delta g$, a regression line is fitted to the point cloud via linear regression for each combination of classifier and splitting function. The tests were conducted with $N_{iter}=10$ iterations.}
  \hypertarget{fig:fc_vs_delta}{}
  \label{fig:nclass}
\end{figure*}

\begin{table*}[t!]
 \caption{P-values from multiple linear regression tests assessing the independence between dataset features and the performance difference between HC and FC, denoted as $\Delta g$. The tests were conducted with $N_{iter}=10$ iterations.}\label{tab:linreg}
 \hypertarget{tab:linreg}{}
  \centering
  \begin{tabular}{lcccccc}
    \toprule
    \multirow{2}{*}{Features} & \multicolumn{3}{c}{\texttt{svm}} & \multicolumn{3}{c}{\texttt{rocket}} \\
    \cmidrule(r){2-4} \cmidrule(r){5-7}\\
    & \texttt{potr} & \texttt{srtr} & \texttt{lsoo} & \texttt{potr} & \texttt{srtr} & \texttt{lsoo}\\  
    \cmidrule(r){2-2} \cmidrule(r){3-3} \cmidrule(r){4-4} \cmidrule(r){5-5} \cmidrule(r){6-6} \cmidrule(r){7-7} \\
\#class&\textbf{0.000}&\textbf{0.000}&\textbf{0.000}&0.165&0.033&0.967\\
FCscore&0.018&0.070&0.133&\textbf{0.000}&\textbf{0.000}&\textbf{0.000}\\
BFD&0.207&0.668&\textbf{0.000}&0.624&0.445&0.893\\
BFC&\textbf{0.000}&0.676&$-$&0.256&0.350&$-$\\
    \bottomrule
  \end{tabular}
  \label{tab:table}
\end{table*}

\begin{table*}[t!]
 \caption{P-values obtained from multiple logistic regression tests, assessing the independence between dataset features and binary improvement situations (i.e., whether there is an improvement or not) when HC is used. The tests were conducted with $N_{iter}=10$ iterations.}\label{tab:logit}
 \hypertarget{tab:logit}{}
  \centering
  \begin{tabular}{lcccccc}
    \toprule
    \multirow{2}{*}{Features} & \multicolumn{3}{c}{\texttt{svm}} & \multicolumn{3}{c}{\texttt{rocket}} \\
    \cmidrule(r){2-4} \cmidrule(r){5-7}\\
    & \texttt{potr} & \texttt{srtr} & \texttt{lsoo} & \texttt{potr} & \texttt{srtr} & \texttt{lsoo}\\  
    \cmidrule(r){2-2} \cmidrule(r){3-3} \cmidrule(r){4-4} \cmidrule(r){5-5} \cmidrule(r){6-6} \cmidrule(r){7-7} \\
\#class&\textbf{0.000}&\textbf{0.000}&\textbf{0.000}&0.881&0.075&0.532\\
FCscore&\textbf{0.003}&\textbf{0.005}&0.023&\textbf{0.000}&\textbf{0.000}&\textbf{0.000}\\
BFD&0.529&0.617&0.603&0.895&0.193&0.592\\
BFC&0.025&0.201&$-$&0.863&0.368&$-$\\
    \bottomrule
  \end{tabular}
  \label{tab:table}
\end{table*}

\end{document}